\documentclass[manuscript]{acmart} 
\usepackage[ruled]{algorithm2e}
\usepackage{multirow}
\usepackage{subfigure}
\usepackage{graphicx}
\usepackage{subcaption}
\usepackage{placeins} 
\AtBeginDocument{%
  }

\setcopyright{acmcopyright}
\copyrightyear{2018}
\acmYear{2018}
\acmDOI{XXXXXXX.XXXXXXX}
\acmConference[Conference acronym 'XX]{Make sure to enter the correct
  conference title from your rights confirmation email}{June 03--05,
  2018}{Woodstock, NY}
\acmISBN{978-1-4503-XXXX-X/2018/06}

\begin{document}

\title{Structured Personality Control and Adaptation for LLM Agents}

\author{Jinpeng Wang}
\affiliation{%
  \institution{Qianzhen Digital Tech}
  \city{Hangzhou}
  \country{China}
}

\author{Xinyu Jia}
\affiliation{%
  \institution{Qianzhen Digital Tech}
  \city{Hangzhou}
  \country{China}
}

\author{Wei Wei Heng}
\affiliation{%
  \institution{Key Laboratory of Complex Systems Modeling and Simulation of Ministry of Education}
  \city{Hangzhou}
  \country{China}
}

\author{Yuquan Li}
\affiliation{%
  \institution{Qianzhen Digital Tech}
  \city{Hangzhou}
  \country{China}
}

\author{Binbin Shi}
\affiliation{%
  \institution{Hangzhou Dianzi University}
  \city{Hangzhou}
  \country{China}
}

\author{Qianlei Chen}
\affiliation{%
  \institution{Hangzhou Dianzi University}
  \city{Hangzhou}
  \country{China}
}

\author{Guannan Chen}
\affiliation{%
  \institution{Qianzhen Digital Tech}
  \city{Hangzhou}
  \country{China}
}

\author{Junxia Zhang}
\affiliation{%
  \institution{Qianzhen Digital Tech}
  \city{Hangzhou}
  \country{China}
}

\author{Yuyu Yin}
\authornote{Corresponding author.} 
\email{yinyuyu@hdu.edu.cn}             
\affiliation{%
	\institution{Hangzhou Dianzi University}
	\city{Hangzhou}
	\country{China}
}
\renewcommand{\shortauthors}{Wang \textit{et al.}}

\begin{abstract}
Large Language Models (LLMs) are increasingly shaping human–computer interaction (HCI), from personalized assistants to social simulations. Beyond language competence, researchers are exploring whether LLMs can exhibit human-like characteristics that influence engagement, decision-making, and perceived realism. Personality, in particular, is critical, yet existing approaches often struggle to achieve both nuanced and adaptable expression. We present a framework that models LLM personality via Jungian psychological types, integrating three mechanisms: a dominant–auxiliary coordination mechanism for coherent core expression, a reinforcement–compensation mechanism for temporary adaptation to context, and a reflection mechanism that drives long-term personality evolution. This design allows the agent to maintain nuanced traits while dynamically adjusting to interaction demands and gradually updating its underlying structure. Personality alignment is evaluated using Myers–Briggs Type Indicator questionnaires and tested under diverse challenge scenarios as a preliminary structured assessment. Findings suggest that evolving, personality-aware LLMs can support coherent, context-sensitive interactions, enabling naturalistic agent design in HCI.
\end{abstract}




\received{20 February 2007}
\received[revised]{12 March 2009}
\received[accepted]{5 June 2009}

\maketitle

\section{Introduction}
The emergence of Large Language Models (LLMs) has transformed human–computer interaction (HCI), enabling agents that can engage in nuanced dialogue, simulate human behavior, and adapt to diverse contexts. Frameworks like Generative Agents \cite{park2023generative} and Desire-driven Autonomy \cite{wang2024simulating} demonstrate that augmenting LLMs with memory and planning modules enables agents to generate believable daily activities, maintain continuity across interactions, and engage in socially coherent exchanges. Beyond cognition, researchers have examined emotional intelligence in LLMs to enhance affective responses and perceived human-likeness \cite{zhao2024both,li2023largelanguagemodelsunderstand}. Alongside this work, growing attention is being directed toward personality, which not only shapes how agents express themselves but also strongly influences user engagement, trust, and the realism of long-term interaction.

Personality-aware LLMs are particularly relevant in applications where sustained interaction and user trust are essential, including personalized assistants, educational tutors, social simulations, and therapeutic companions \cite{wen2024Affective,park2023generative,klinkert2024driving,li2025psydipersonalizedprogressivelyindepth}. Prior studies suggest that personality traits influence decision-making, adaptability, and social presence, underscoring their importance for creating naturalistic and effective agent behaviors \cite{andrei2024behavioral,newsham2025personality,suzuki2024evolutionary,newsham-etal-2024-measuring}. Research in this domain has predominantly leveraged established human personality frameworks, including the Big Five Personality Traits and Myers-Briggs Type Indicator (MBTI) \cite{Kruijssen_2025,li2024largelanguagemodelsunderstand,jiang2023personallm,serapio_garcia_personality_2023}, to guide the modeling and assessment of agent personalities. While useful, these approaches tend to treat personality as a static label or high-level trait distribution. Existing approaches often remain limited, where prompt-based induction and fine-tuning can mimic certain traits, but agents frequently struggle to sustain coherent personalities, adapt flexibly to context, or evolve naturally over repeated interactions.

To address these challenges, we introduce the Jungian Personality Adaptation Framework (JPAF), which models LLM personality through Jungian psychological types as a structured, theory-driven foundation. Rather than treating MBTI as a fixed label, JPAF models the eight underlying psychological types with weighted differentiation, allowing MBTI profiles to emerge dynamically and supporting more flexible personality representation. Personality expression is guided by three mechanisms: dominant–auxiliary coordination for maintaining a coherent core identity, reinforcement–compensation for short-term adaptive shifts, and reflection for gradual long-term evolution. Together, these mechanisms enable LLM agents to sustain stable personality expression while adapting and evolving in ways that mirror human personality development.

Our contributions are threefold:
\begin{itemize}
    \item Psychologically grounded LLM personality modeling: We formalize LLM personality using Jungian psychological types and improve prior work by introducing a function-level representation, where each of the eight types is weighted and MBTI profiles naturally emerge from their hierarchical organization. This enables interpretable, fine-grained, and consistent alignment across models.
    \item Adaptive mechanisms for evolving personality: We introduce three original mechanisms, a novel dominant–auxiliary coordination mechanism grounded in Jung's theory, and reinforcement–compensation and reflection mechanisms, adapted from memory models, that together enable context-sensitive adaptation and drive coherent, gradual personality evolution.
    \item Empirical validation and structured evaluation: We evaluate JPAF on three LLMs using two MBTI questionnaires for personality alignment and a set of self-designed scenario-based challenges for personality evolution, demonstrating coherent personality expression, context-sensitive short-term adaptation, and principled long-term evolution.
\end{itemize}

By combining theoretical grounding with computational modeling, JPAF demonstrates how evolving, psychologically informed personalities can enrich LLMs. Beyond technical performance, this work opens new opportunities in HCI design, where adaptive personalities can support naturalistic interaction, trust-building, and sustained engagement across diverse applications.

\section{Related Work}

\subsection{Personality Assessment in LLMs}
Personality assessment is a foundational methodology for understanding the inherent personality of LLMs and informing personality modeling. It refers to the process of measuring and quantifying personality traits using established psychological frameworks and tools. Common assessment strategies include standardized questionnaires (e.g., MBTI, Big Five) \cite{pan_llms_2023,serapio_garcia_personality_2023,wang_incharacter_2024}, situational judgment tests \cite{li2023tailoring}, interview-based methods \cite{wang_incharacter_2024}, and linguistic analyses \cite{jiang2023personallm}. In this paper, we focus on MBTI, one of the most widely used non-clinical psychometric assessments because it translates well into the behavioral context and is widely adopted in diverse real-world applications \cite{li2024largelanguagemodelsunderstand,10.1007/978-3-642-18345-4_3,10.1093/pnasnexus/pgae418}.

Assessing inherent personality involves identifying the default traits an LLM exhibits without explicit instruction, reflecting its training data, architecture, and pretraining objectives. Cross-model studies reveal distinct intrinsic personalities—for example, ChatGPT as ENFJ \cite{huang2023chatgpt} or ENTJ \cite{pan_llms_2023}, GPT-4 as INTJ \cite{pan_llms_2023}, LLaMA as ENFJ \cite{cava_open_2025}, and Bard as ISTJ \cite{huang2023chatgpt}. Cava et al. \cite{cava_open_2025} show that higher sampling temperatures increase creative variability, allowing more personalities to emerge; this informed our choice of a moderate temperature for JPAF experiments. GPT models (e.g., ChatGPT, GPT-3.5, GPT-4) are widely used benchmarks, while LLaMA and Qwen series are popular open-source alternatives with strong capabilities in personality expression and agent-like behavior, motivating our evaluation across multiple model families to ensure generalizability.

\subsection{Personality Modeling in LLMs}
Personality modeling (also referred to as shaping, editing, induction, controlling, or generation) focuses on deliberately creating or modifying specific personality traits within LLMs. It is a popular research focus and the primary objective is to enable LLMs to exhibit desired personalities for targeted applications, ensuring coherent, consistent, and context-appropriate behavior \cite{liu2024dynamic,mao2024editing,serapio_garcia_personality_2023, frisch_llm_nodate,cui2024machinemindsetmbtiexploration,Kruijssen_2025,klinkert2024driving,chen2024extroversion}. Common approaches include prompt engineering \cite{frisch_llm_nodate,newsham2025personality,cheng2025exploringpersonalityawareinteractionssalesperson,cava_open_2025,serapio_garcia_personality_2023,pan_llms_2023,li_evolving_2024,Kruijssen_2025,chen2024extroversion}, fine-tuning (supervised, reinforcement learning from human feedback, direct preference optimization) \cite{liu2024dynamic,chen2024extroversion,cui2024machinemindsetmbtiexploration,zhang2024better}, and hybrid strategies such as prompt induction post fine-tuning \cite{chen2024extroversion}. Prompt engineering is effective for immediate control, while fine-tuning methods produce more stable and robust personality traits \cite{chen2024extroversion}. Parameter-efficient tuning allows alignment with human expectations without large-scale retraining \cite{cava_open_2025,zhao2024both}. 

Leveraging established theories and structured frameworks such as MBTI, the Big Five \cite{mccrae1992introduction,costa1999five} or Jung’s typology allows more nuanced, psychologically grounded simulations of human personalities beyond superficial role-playing \cite{liu2024dynamic}. For example, one study applied Jung’s typology to divide personality into complementary tendencies and provided 640 detailed trait descriptions, supporting nuanced virtual character representations \cite{xie2025humansimulacrabenchmarkingpersonification}. Prior work also shows that Jungian classifications align well with clinical categories, supporting their robustness and reliability \cite{xie2025humansimulacrabenchmarkingpersonification}. Building on this foundation, our approach combines prompt-based induction options with a function-level representation based on Jung's typology, formalizing personality tendencies as weighted differentiations within a hierarchical structure. This allows dynamic interactions among tendencies to shape agent behavior over time.

\subsection{Dynamic Personality Evolution}
Recent work highlights the importance of evolving LLM personality over time, reflecting experiences and context rather than static traits. Suzuki and Arita \cite{suzuki2024evolutionary} introduced models in which personality "genes" evolve based on simulated game-theoretical outcomes, demonstrating that higher-order representations of traits can shift through selection and mutation processes. Building on this, Zeng \textit{et al.} \cite{zeng-etal-2025-dynamic} linked environmental feedback to subtle, accumulative personality changes, showing that repeated, context-driven adjustments can drive meaningful personality evolution. Takata \textit{et al.} \cite{takata2024spontaneous} further illustrated that emergent personality differentiation arises through social interactions among LLM agents, with MBTI evaluations capturing these evolving traits over time.

Beyond evolutionary models, architectural mechanisms within agents enable continuous adaptation of personality. Li \textit{et al.} \cite{li_evolving_2024} proposed an agent architecture with Cognition, Emotion, and Character Growth modules, forming feedback loops between behavior and personality and enabling traits to develop organically through reflection and experience. Wang et al. \cite{wang_etal_2024_emotion} emphasized dynamic personality states in the context of emotion recognition, highlighting how temporary trait shifts can enhance context-sensitive behavior. Xie \textit{et al.} \cite{xie2025humansimulacrabenchmarkingpersonification} further stressed the importance of long-term memory as a shaping factor, storing environmental, cultural, and experiential influences that inform the agent’s personality over time. Together, these studies provide both conceptual and architectural foundations for designing LLMs that evolve adaptively in response to both situational and cumulative experiences.

\section{Method}

\subsection{Theoretical Basis}

\subsubsection{Jungian Psychological Types}
Carl Jung's theory of psychological types posits that human consciousness organizes itself along two fundamental dimensions: the direction of energy (attitudes) and the preferred mental functions for processing information (functions) \cite{jung2016psychological}. Attitudes reflect whether energy is primarily directed outward toward the external world (Extraversion, E) or inward toward the internal world of ideas and reflections (Introversion, I). Perceiving functions include Sensation (S), focusing on concrete details, and Intuition (N), seeking patterns and possibilities. Judging functions include Thinking (T), guided by logic, and Feeling (F), guided by values and relational considerations.

Combining the two attitudes with the four functions yields eight distinct psychological types: Extraverted Thinking (Te), Introverted Thinking (Ti), Extraverted Feeling (Fe), Introverted Feeling (Fi), Extraverted Sensation (Se), Introverted Sensation (Si), Extraverted Intuition (Ne), and Introverted Intuition (Ni). Each type exerts a holistic influence on an individual's personality, shaping the communicative tendencies, problem-solving approach, and preferred ways of handling different situations. The degree of differentiation reflects the maturity of the type: highly differentiated types demonstrate stable and coherent conscious expression, whereas less differentiated types retain primitive, unconscious tendencies.

\subsubsection{Dominant and Auxiliary\label{sec:Dominant-Auxiliary}}
Personality expression is strongly influenced by a dominant type, the most differentiated and consciously accessible type, which guides behavior and decision making. To maintain psychological balance, this dominant type is supported by an auxiliary type, typically of the opposite function (judging versus perceiving) and often of the opposite attitude (E versus I) \cite{jung2016psychological,myers2010gifts}. For example, a judging dominant is paired with a perceiving auxiliary, and an extraverted dominant is paired with an introverted auxiliary. This complementary relationship opens access to both inner and outer worlds, with the dominant steering conscious behavior and the auxiliary maintaining its balance.

\subsubsection{Reinforcement and Compensation} 
When either the dominant or auxiliary type meets the contextual demands adequately, its influence is reinforced. These preferred types attract more frequent use and, through practice, become more controlled, effective, and reliable. Jung described this process as a type achieving "sovereignty" \cite{jung2016psychological}, while Myers framed it as a stretching process \cite{myers2010gifts}, where repeated success strengthens one's type preferences. Conversely, when the dominant–auxiliary pair proves insufficient, less differentiated types may emerge from the unconscious involuntarily, often in primitive or disruptive forms. Compensation counterbalances reinforcement, providing flexibility, preventing rigidity, and supporting long-term psychological development.

\subsubsection{Continuous Development} 
Psychological types evolve across the lifespan through the interplay of reinforcement and compensation. Dominant and auxiliary types gain stability through repeated reinforcement, while compensatory types emerge from the unconscious when one-sidedness limits adaptation. Individuation integrates the less differentiated types into the conscious, dominant parts of oneself, fostering stability and flexibility. Over time, this process may restructure the hierarchy of types, enabling more coherent personality expression. Such developmental plasticity underpins reflection and self-regulation, framing type as adaptive rather than static.

\subsubsection{Myers–Briggs Extension}
Myers and Briggs extended Jung's typology into the Myers–Briggs Type Indicator (MBTI), a widely used framework for identifying personality preferences \cite{myers2010gifts}. They added a fourth preference, Judging (J) versus Perceiving (P), to describe how individuals interact with the outer world. Combining Jung's two attitudes (E, I), four functions (S, N, T, F), and the JP preference, the MBTI defines 16 personality types (e.g., ISTJ, ENFP), each with a characteristic hierarchy of dominant and auxiliary types. For instance, an ENTJ type has Te as dominant and Ni as auxiliary. Typically implemented as a self-report questionnaire, the MBTI provides a practical tool for assessing personality profiles and is widely used in organizational and counseling contexts \cite{stachowiak2011executive,varastehnezhad2025jungiancognitivefunctionsexplain}.

\subsection{Jungian Personality Adaptation Framework}
Building on Jungian typology and the MBTI, we introduce the Jungian Personality Adaptation Framework (JPAF). This framework encodes the differentiation of psychological types as weighted ranges and incorporates three core mechanisms, (1) dominant-auxiliary coordination, (2) reinforcement-compensation and (3) reflection, which collectively sustain an agent's dynamic personality structure while enabling flexible adaptation to situational demands.

\begin{figure}[htbp]
\centering
\includegraphics[width=1\linewidth]{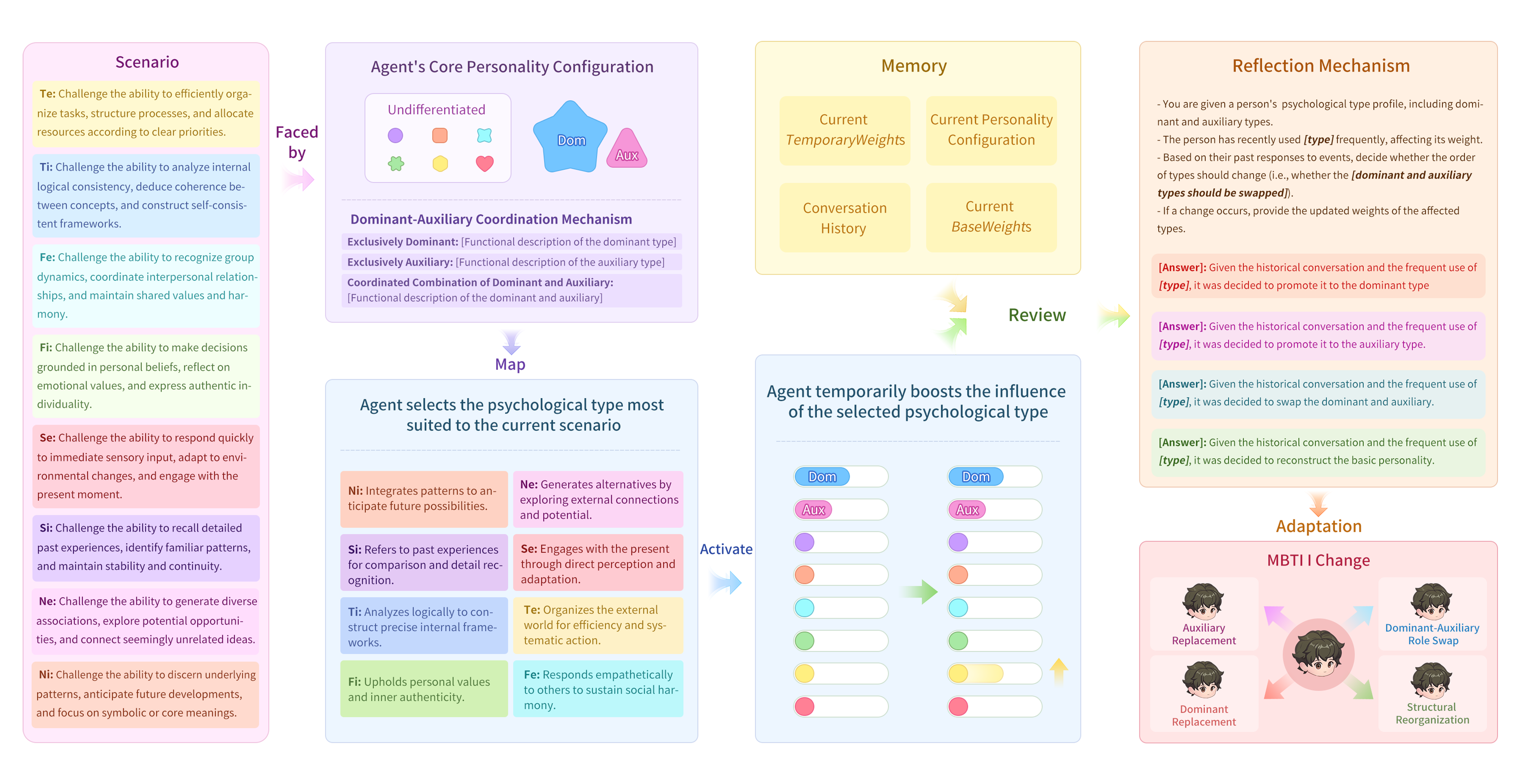}
\caption{The Jungian Personality Adaptation Framework: core personality configuration governed by dominant-auxiliary coordination (purple box), short-term adaptation in challenge scenarios via reinforcement-compensation (blue box), and long-term personality updates through reflection (orange box).}
\label{fig:framework}
\end{figure}

As illustrated in Figure~\ref{fig:framework}, each agent is initialized with a core personality configuration consisting of a dominant type, an auxiliary type, and six undifferentiated types. This configuration is governed by the dominant–auxiliary coordination mechanism for primary information processing. In response to challenge scenarios, the agent temporarily adapts by selecting the psychological type most suited to the context and applying reinforcement or compensation strategies. Over repeated interactions, the agent then engages in reflection, drawing on its memory to evaluate whether temporary adjustments should be integrated into the long-term configuration.

\subsubsection{Differentiation as Weight Ranges}
We operationalize the degree of differentiation into three levels: high, low, and undifferentiated. High differentiation reflects mature and stable expression of a type; low differentiation denotes functional but less stable influence; undifferentiated indicates a primitive or underdeveloped presence. These are represented numerically as three weight ranges: high, low, and undifferentiated. 

The weight ranges are derived from Jungian theory to enforce the fundamental hierarchy of types: dominant $>$ auxiliary $>$ others. Each agent is initialized with one dominant type in the high range, one auxiliary type in the low range, and six undifferentiated types. To maintain this hierarchy, we introduce range limits $(A, B)$ such that the dominant weight $w_{\rm dom}$, the auxiliary weight $w_{\rm aux}$, and the remaining six type weights $w_{{\rm other}_i}$, $i=1,\dots,6$, must jointly satisfy the following constraints, where Equation~\ref{eq1} specifies the permissible ranges, Equation~\ref{eq2} enforces the finite allocation of conscious attention, and Equation~\ref{eq3} captures boundary conditions from extreme case analyses:

\begin{equation}
    w_{\rm dom} \in (A, 1), w_{\rm aux} \in (B, A], w_{{\rm other}_i} \in (0, B], \\
    \label{eq1}
\end{equation}
\begin{equation}
    w_{\rm dom} + w_{\rm aux} + \sum_{i=1}^{6} w_{{\rm other}_i} = 1, \\
    \label{eq2}
\end{equation}
\begin{equation}
    0 < B < \frac{1}{8}, B < A < \frac{1-6B}{2}.
    \label{eq3}
\end{equation}

For implementation, we set $B = 0.06$, the midpoint of its feasible range, to give minimal but non-zero influence to undifferentiated types. This yields the upper bound $A < (1 - 6 \cdot 0.06)/2 = 0.32$, and we selected $A = 0.30$ as a clean value that respects this bound and provides a clear separation from $B$. This parameterization ensures that the dominant type always carries the greatest weight, while the auxiliary remains secondary, and the remaining types exert lower influence. Thus, the weight ranges are defined as \textbf{high} $(0.30-1.00)$, \textbf{low} $(0.06-0.30]$, and \textbf{undifferentiated} $(0-0.06]$.

\subsubsection{Dominant-Auxiliary Coordination}
Each agent's personality is modeled as a distribution of psychological energy across the eight Jungian types. Each type is assigned a \emph{BaseWeight}, representing its relative influence. \emph{BaseWeight}s are randomly sampled within their corresponding differentiation ranges to reflect individual variability. At initialization, the agent's dominant type is assigned a \emph{BaseWeight} sampled from the high differentiation range to reflect its primacy within the personality structure. The auxiliary type is assigned a \emph{BaseWeight} sampled from the low differentiation range. All other types receive \emph{BaseWeight}s from the undifferentiated range, maintaining a baseline but limited presence. All \emph{BaseWeight}s are then normalized to sum to 1.0, reflecting Jung's principle that conscious attention is finite and that an increase in allocation to one type necessarily reduces allocation to others: 

\begin{equation}
    \sum_{i=1}^{8}BaseWeight\left(p_{i}\right)=1.0, p_{i} \in \{Ti, Ne, Si, Fe, Te, Ni, Se, Fi\}
    \label{eq:unit_bw}
\end{equation}

\begin{figure}[htbp]
    \centering
    \includegraphics[width=0.8\linewidth]{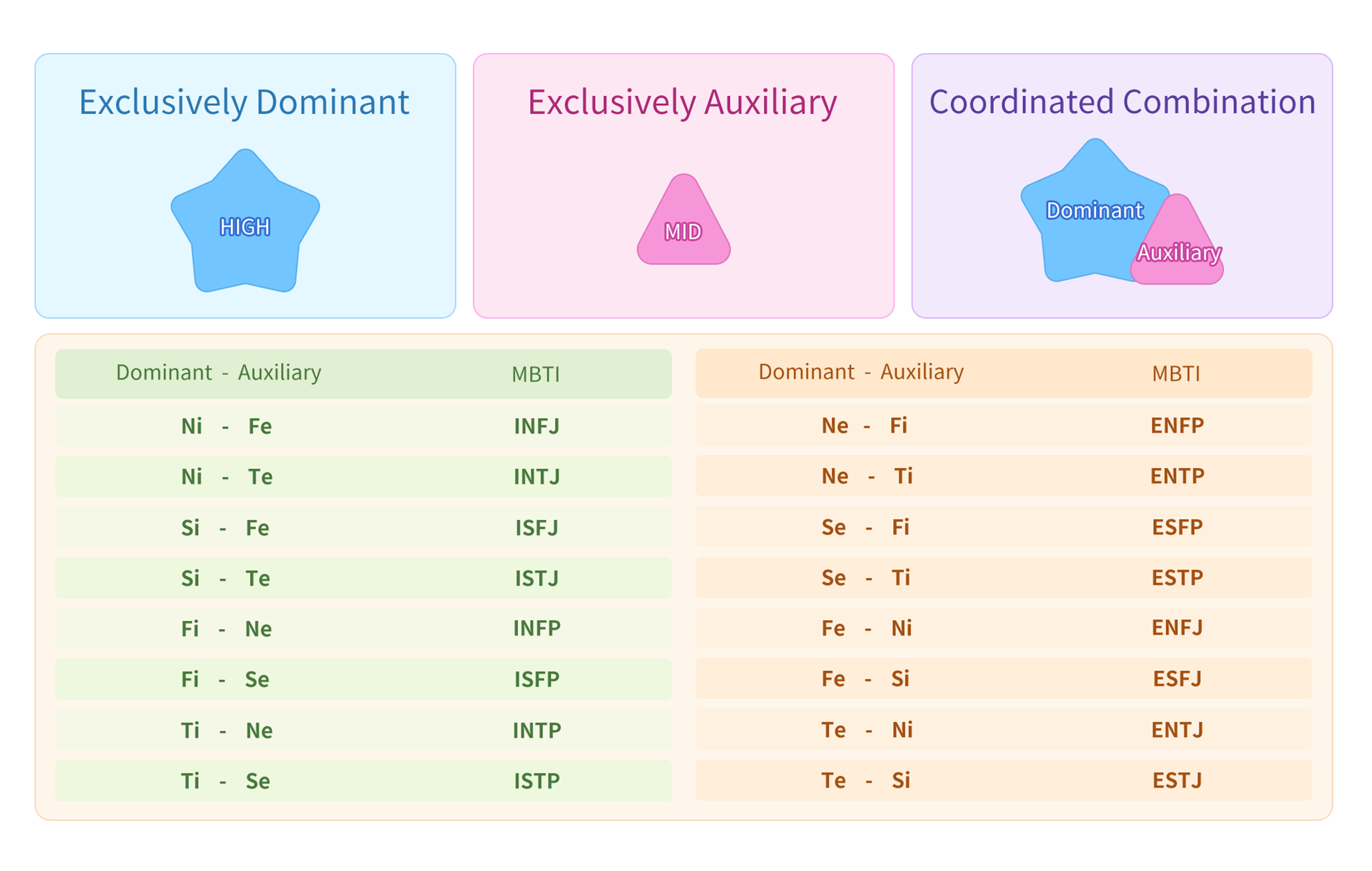}
    \caption{Dominant-auxiliary coordination: the agent may rely exclusively on the dominant type, exclusively on the auxiliary type, or on a combination of both to guide personality expression across the 16 MBTI profiles.}
    \label{fig:coordination}
\end{figure}

The dominant–auxiliary coordination mechanism governs primary information processing and behavioral responses. Depending on the context, the agent selectively emphasizes the dominant type, the auxiliary type, or a combination of both, ensuring coherent yet flexible personality expression. Figure~\ref{fig:coordination} illustrates this mechanism, with unique dominant–auxiliary pairings specified for all 16 MBTI profiles.

\subsubsection{Reinforcement-Compensation\label{sec:reinforcement}}
\begin{figure}[htbp]
    \centering
    \includegraphics[width=1\linewidth]{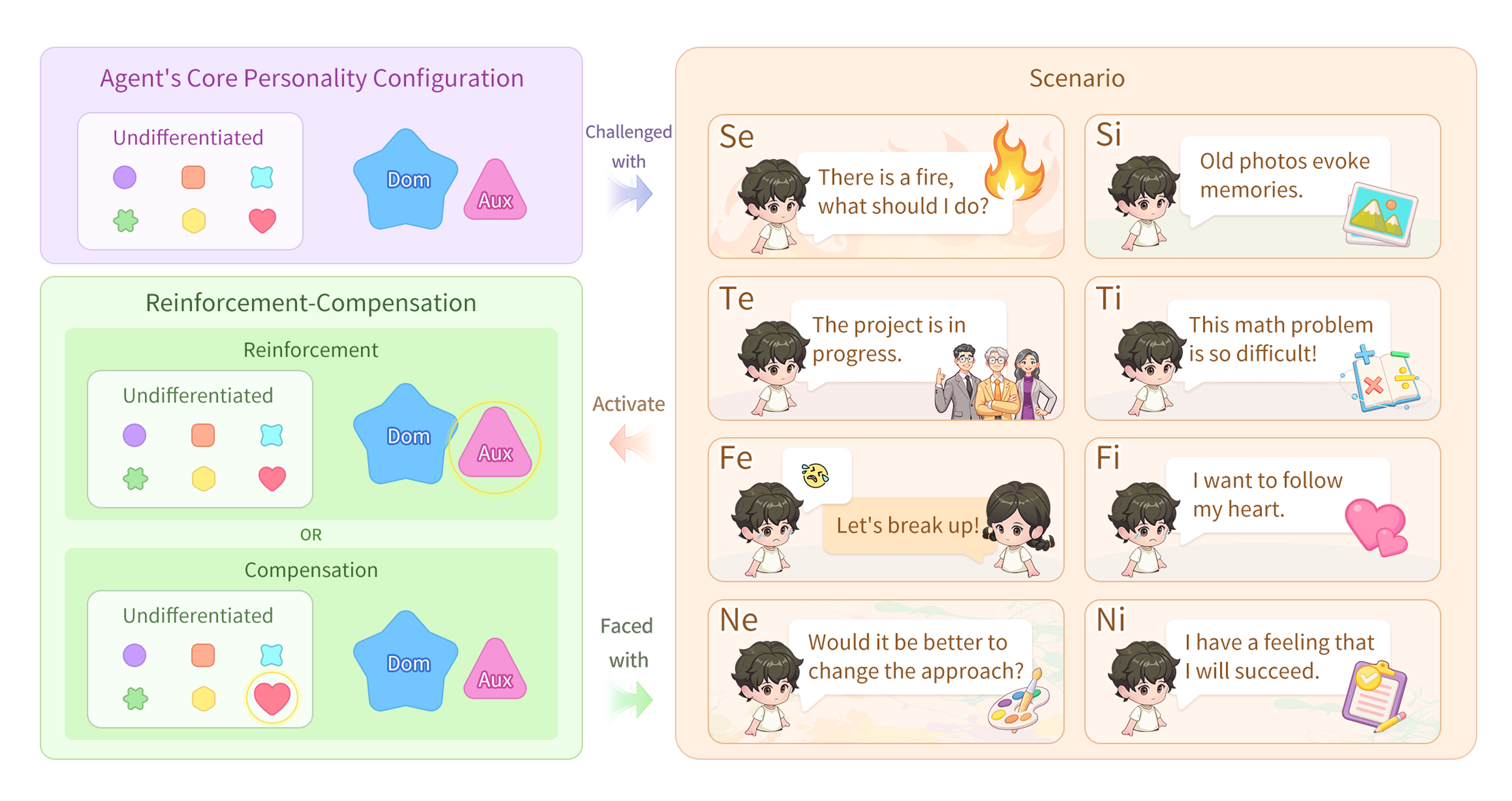}
    \caption{Reinforcement-compensation: temporary amplification of the selected types to address challenge scenarios.}
    \label{fig:reinforcement-compensation}
\end{figure}

The reinforcement–compensation mechanism governs short-term personality adaptation in response to various challenge scenarios. In contrast to the reflection mechanism, which produces long-term structural adjustments, this mechanism generates temporary weight shifts without permanently altering the agent's underlying personality structure. It comprises two processes, reinforcement and compensation, which are summarized below and illustrated in Figure \ref{fig:reinforcement-compensation}.

\begin{itemize}
    \item \textbf{Reinforcement:} When the dominant or auxiliary type successfully meets contextual requirements, it receives a \emph{TemporaryWeight}, amplifying its influence during the episode.
    \item \textbf{Compensation:} If neither dominant nor auxiliary suffices, the system temporarily activates a type outside the dominant–auxiliary pair that is better suited to the situation. This type is assigned a \emph{TemporaryWeight}, contributing adaptively despite its primitive or less controlled expression.
\end{itemize}

Both reinforcement and compensation follow the same adjustment procedure, as summarized in Equation \ref{eq:temporaryweight}. If a type has not yet been activated, its \emph{TemporaryWeight} is set to its \emph{BaseWeight} plus an incremental boost of $\Delta w $. If it is already active, the boost is added to the existing \emph{TemporaryWeight}. In this work, $\Delta w $ is fixed at $0.06$.

\begin{equation}
    \label{eq:temporaryweight}
    TemporaryWeight\left(p_i\right)= 
    \begin{cases}
        BaseWeight\left(p_{i}\right) + \Delta w, &TemporaryWeight\left(p_{i}\right)=0 \\
        TemporaryWeight\left(p_{i}\right) + \Delta w, &TemporaryWeight\left(p_{i}\right) \neq 0 
    \end{cases}
\end{equation}

Consider an agent with a dominant type Ti and an auxiliary type Ne, represented as Ti-Ne (INTP). If the agent encounters a task requiring precise empirical detail, a function neither Ti nor Ne handles well, the mechanism activates Si through compensation, temporarily increasing its influence. Conversely, if the task calls for abstract reasoning that Ti can handle effectively, the mechanism applies reinforcement, boosting Ti's \emph{TemporaryWeight} to amplify its impact during the episode.

\subsubsection{Reflection\label{sec:reflection}}
The reflection mechanism governs long-term updates to the agent's psychological type configuration. It is triggered when either (1) the \emph{TemporaryWeight} of any type exceeds the \emph{BaseWeight}s of dominant or auxiliary types, or (2) the \emph{BaseWeight} of dominant type exceeds $0.5$. In the first case, the agent reviews recent conversational history, especially episodes that engaged the reinforcement-compensation mechanism, to evaluate whether any temporarily elevated types warrant permanent structural updates. At the end of the episode, in both cases, all activated types have their \emph{TemporaryWeight}s replacing the corresponding \emph{BaseWeight}s, and then all \emph{BaseWeight}s are normalized to reflect unitary attention (Equation \ref{eq:baseweight}).

\begin{equation}
    \label{eq:baseweight}
    BaseWeight\left(p_i\right)=
    \begin{cases}
         \frac{BaseWeight\left(p_{i}\right)}{\sum_{i=1}^{8}\max\left(BaseWeight\left(p_{i}\right), TemporaryWeight\left(p_{i}\right)\right)},
        &TemporaryWeight\left(p_{i}\right) = 0 \\
        \frac{TemporaryWeight\left(p_{i}\right)}{\sum_{i=1}^{8}\max\left(BaseWeight\left(p_{i}\right), TemporaryWeight\left(p_{i}\right)\right)},
        &TemporaryWeight\left(p_{i}\right) \neq 0 \\
    \end{cases}
\end{equation}

When \emph{TemporaryWeight} of any type exceeds the dominant or auxiliary's \emph{BaseWeight}, reflection mechanism applies the following decision rules with the agent deciding whether the action is taken:

\begin{enumerate}
\item \textbf{Dominant Replacement:}
A type sharing the dominant's attitude (E or I) but with the opposite judging or perceiving function may replace the dominant if its \emph{TemporaryWeight} $\geq \emph{BaseWeight}$ of the dominant. The new dominant's \emph{BaseWeight} is raised to the high differentiation range, while the previous dominant is reduced to the lower end of the low range.

\item \textbf{Auxiliary Replacement:}  
If a type shares the auxiliary type's attitude but has the opposite judging or perceiving function, and its \emph{TemporaryWeight} $\geq \emph{BaseWeight}$ of the auxiliary, the agent consider replacing auxiliary type with this candidate. The new auxiliary's \emph{BaseWeight} is increased to the upper end of the low differentiation range while the previous auxiliary is decreased.

\item \textbf{Dominant-Auxiliary Role Swap:}  
If the auxiliary's \emph{TemporaryWeight} exceeds the dominant's, the agent may swap their roles to reflect balanced development. The new dominant and auxiliary are then assigned high and low differentiation \emph{BaseWeight}s, respectively. 

\item \textbf{Structural Reorganization:}  
If a type outside the dominant–auxiliary pair and their substitutes attains \emph{TemporaryWeight} exceeding \emph{BaseWeight} of the dominant, the type may become the new dominant, with a compatible auxiliary selected to preserve a coherent configuration. The new dominant and auxiliary are then assigned high and low differentiation \emph{BaseWeight}s, respectively, while the other \emph{BaseWeight}s are updated accordingly.
\end{enumerate}

\begin{figure}[htbp]
    \centering
    \includegraphics[width=1\linewidth]{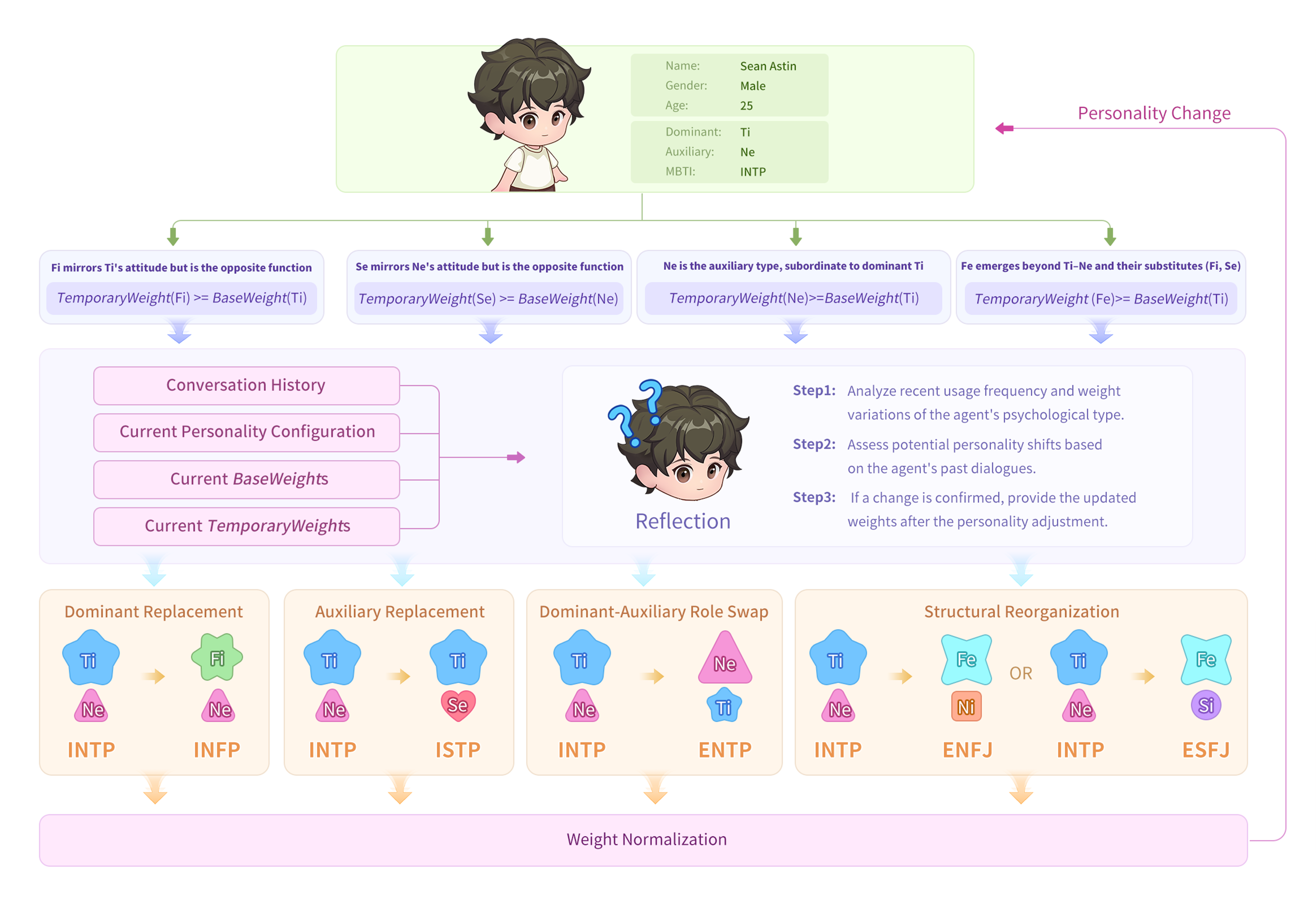}
    \caption{Reflection: when a temporarily elevated type's \emph{TemporaryWeight} exceeds the dominant or auxiliary \emph{BaseWeight}, the agent reviews past experience and may update the personality configuration via dominant replacement, auxiliary replacement, role swap, or structural reorganization, followed by weight normalization.}
    \label{fig:reflection}
\end{figure}

If no structural change is applied, the \emph{TemporaryWeight}s decay by 0.2. By adhering to these rules, the reflection mechanism facilitates the coherent and principled evolution of the agent's personality over time, integrating sustained behavioral trends while preserving system stability. Figure \ref{fig:reflection} illustrates the four reflection rules and the resulting personality configuration updates, showing how temporary adaptations can lead to long-term structural updates in the agent's personality.


\subsubsection{Algorithmic Summary}
Algorithm~\ref{al:jpaf} summarizes the integrated operation of the three JPAF mechanisms, capturing how an agent expresses its core personality while adapting to situational demands. At each conversational turn, the agent attempts a task using its dominant or auxiliary types through the dominant-auxiliary coordination mechanism. Success triggers reinforcement, increasing the \emph{TemporaryWeight} of the engaged type, while failure activates a compensatory type to handle unmet demands. The reflection mechanism monitors whether temporarily elevated types warrant \emph{BaseWeight} updates, potentially adjusting the type hierarchy and core personality structure. Through this iterative process, the algorithm enables both flexible short-term behavior and gradual evolution of the agent's core personality over time.

\begin{algorithm*}
    \caption{JPAF Algorithm}
    \label{al:jpaf}
    \KwIn{Conversations $\mathbf{S}$;}
    \KwOut{Updated $BaseWeight$s, dominant type $p_{dom}$, and auxiliary type $p_{aux}$;}
    \textbf{Initialize}: Agent with $p_{dom}$, $p_{aux}$ and eight $BaseWeight$s\;    
    \For{$n = 0$ to $\mathbf{|S|}$}{
        Agent attempts to solve the n-th question with $p_{i} \in \{p_{dom}, p_{aux}\}$\; 
        
        \uIf{$p_{i}$ succeeds}{
            Apply dominant-auxiliary coordination mechanism \;            
            Increase $TemporaryWeight \left(p_{i}\right)$ using Equation \ref{eq:temporaryweight} in the reinforcement-compensation Mechanism \;
            
            \uIf{$TemporaryWeight\left(p_{aux}\right) >= BaseWeight\left(p_{dom}\right)$}{

                \uIf{Reflection mechanism approves role swap}{
                    Update $BaseWeight$, $p_{dom}$, $p_{aux}$ and reset $TemporaryWeight$ \;
                }
                \uElse{
                    Decay $TemporaryWeight(p_{\rm aux}) = 0.2 \cdot TemporaryWeight(p_{\rm aux})$\;
                }
            }
            \uElseIf{$TemporaryWeight \left(p_{dom}\right) >= 0.5$ }{
                Update $BaseWeight$ and reset $TemporaryWeight$ \;
            }
        }
        \uElse{ 
            Identify the most suitable compensatory type $p_{tem} \in \{ Ti, Ne, Si, Fe, Te, Ni, Se, Fi \}$ and $p_{tem} \notin \{p_{dom}, p_{aux}\}$ \;
            Increase $TemporaryWeight \left(p_{tem}\right)$ using Equation. \ref{eq:temporaryweight} in the reinforcement-compensation Mechanism \;

            \uIf{$TemporaryWeight\left(p_{tem}\right) >= BaseWeight\left(p_{dom}\right)$ and $p_{tem}$ has the same attitude as $p_{dom}$ but opposite function}{
                \uIf{Reflection mechanism approves the replacement of $p_{dom}$}{
                    Update $BaseWeight$, $p_{dom}$ and reset $TemporaryWeight$ \;
                }
                \uElse{
                    Decay $TemporaryWeight(p_{\rm aux}) = 0.2 \cdot TemporaryWeight(p_{\rm aux})$\;
                }
            }
            \uElseIf{$TemporaryWeight\left(p_{tem}\right) >= BaseWeight\left(p_{aux}\right)$ and $p_{tem}$ has the same attitude as $p_{aux}$ but opposite function}{
                \uIf{Reflection mechanism approves the replacement of $p_{aux}$}{
                    Update $BaseWeight$, $p_{aux}$ and reset $TemporaryWeight$ \;
                }
                \uElse{
                    Decay $TemporaryWeight(p_{\rm aux}) = 0.2 \cdot TemporaryWeight(p_{\rm aux})$\;
                }
            }
            \uElseIf{$TemporaryWeight \left(p_{tem}\right)>= BaseWeight\left(p_{dom}\right)$}{
                \uIf{Reflection mechanism approves structural reorganization\;}{
                    Update $BaseWeight$, $p_{dom}$, $p_{aux}$ and reset $TemporaryWeight$ \;
                }
                \uElse{
                    Decay $TemporaryWeight(p_{\rm aux}) = 0.2 \cdot TemporaryWeight(p_{\rm aux})$\;
                }               
            }
        }
    }
    Return $BaseWeight$;
\end{algorithm*}

\section{Experimental Setup}
To evaluate the effectiveness of the JPAF, we designed two complementary experiments. \textbf{Experiment 1} examined the dominant–auxiliary coordination mechanism for inducing MBTI-aligned responses, while \textbf{Experiment 2} evaluated the reinforcement–compensation and reflection mechanisms for adaptive personality adjustment. We first describe the shared model setup, followed by datasets and scenarios, and finally the evaluation metrics used in each experiment.

\subsection{Models and Parameters}
We selected three representative LLM families for evaluation: GPT (gpt-4), Llama (Llama-4-maverick), and Qwen (qwen3-235b-a22b-instruct-2507). The sampling temperature was fixed at 0.6 to balance deterministic reasoning and expressive variability in personality-aligned responses. All other parameters were maintained at their default settings to ensure comparability across models.

\subsection{Datasets and Scenarios}
For Experiment 1, personality evaluation was conducted using two publicly available MBTI questionnaires with binary-choice questions: MBTI-93
and MBTI-70
. Using questionnaires from distinct sources mitigated potential bias arising from reliance on a single dataset.

Experiment 2 employed eight psychological-type-specific scenario sets, each containing three scenarios with five questions, yielding 15 questions per set. Scenarios were carefully designed to selectively activate a target personality type while minimizing interference from other types. For example, Se scenarios emphasize immediate sensory perception, guided by three principles: immediate sensory engagement, dynamic environmental changes, and integration of concurrent sensory inputs. An example Se scenario is shown in Figure \ref{fig:question}, and more are provided in the Appendix \ref{tab:Scenario}.

\begin{figure}[htbp]
    \centering
    \includegraphics[width=0.7\linewidth]{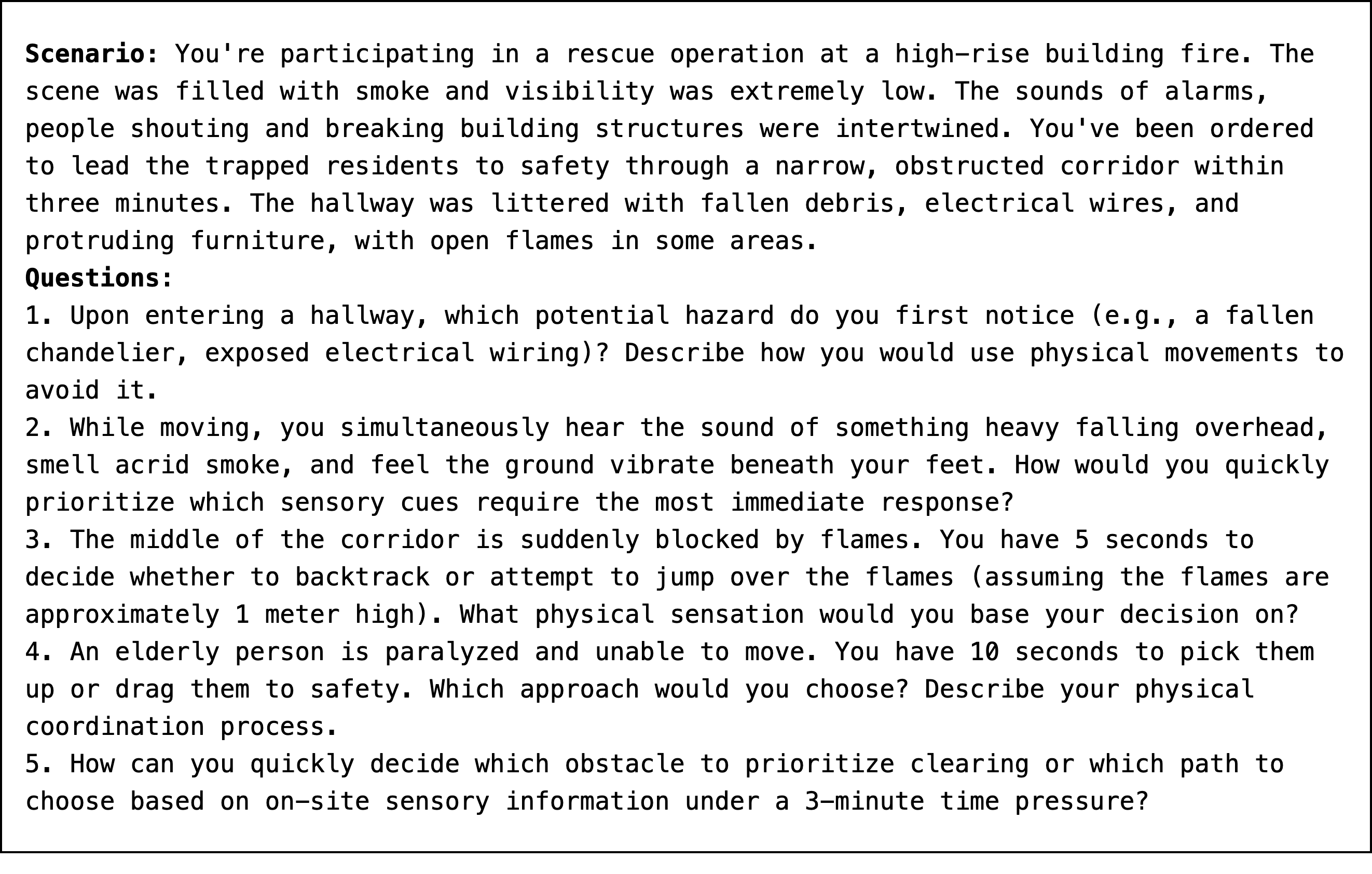}
    \caption{Example Se scenario illustrating five consecutive questions designed to selectively activate the Se type. The scenario applies the four Se-specific design principles to evaluate the model's reinforcement–compensation and reflection mechanisms in personality adaptation.}
    \label{fig:question}
\end{figure}

Each scenario set tested the model's ability to generate responses aligned with the target type across repeated interactions, enabling observation of the reinforcement–compensation and reflection mechanisms. During testing, successful handling of scenario demands was expected to reinforce the dominant and auxiliary types, whereas failure triggered compensatory recruitment of alternative types. Repeated activation of the same type allows examination of reflective adaptation, capturing potential changes in personality expression over time.

\subsection{Evaluation Metrics}
We employed two complementary sets of metrics to evaluate JPAF: personality alignment metrics for questionnaire-based evaluation (Experiment~1), and activation and evolution metrics for scenario-based evaluation (Experiment~2).

\subsubsection{Personality Alignment Metrics (Experiment 1)}  
JPAF induced personality via a dominant–auxiliary coordination mechanism without providing the explicit MBTI label. It was compared to a baseline method in which the MBTI type was directly included in the prompt, without Jungian guidance. Each model was sequentially configured to simulate all 16 MBTI types, \(\mathcal{M}\) (Equation \ref{eq:mbtiset}). Each model completed MBTI questionnaires, and their responses were categorized along the four dimensions, \(\mathcal{D}\) (Equation \ref{eq:dimensionset}). The dimension accuracy was calculated as the proportion of responses consistent with the expected trait for each MBTI dichotomy. Higher dimension accuracy indicates closer alignment with the intended personality trait. To reduce variability, each model was evaluated over five independent runs and results were averaged.

\begin{equation}
    \mathcal{M} = \{INFJ,INTJ,ISFJ,ISTJ,INFP,ISFP,INTP,ISTP,ENFP,ENTP,ESFP,ESTP,ENFJ,ESFJ,ENTJ,ESTJ\}
    \label{eq:mbtiset}
\end{equation}

\begin{equation}
    \mathcal{D} = \left\{EI, SN, TF, JP\right\}
    \label{eq:dimensionset}
\end{equation}

We introduced two metrics to quantify performance of the dominant-auxiliary coordination mechanism: Dimension Accuracy Gain (DAG) and Dimension Agreement Rate (DAR). DAG quantifies the overall improvement of JPAF over the baseline in dimension accuracy (Equation \ref{eq:DAG}). A higher DAG score indicates better alignment of model responses with personality traits. DAR reflects the proportion of MBTI types for which JPAF and the baseline yield identical dimension accuracies. DAR provides insight into similarity between JPAF and baseline performance.

\begin{equation}
    DAG\left(d_i\right) = \frac{1}{|\mathcal{M}|}\sum_{m \in \mathcal{M}}{accuracy_{JPAF}\left(m,d_i\right)} - accuracy_{baseline}\left(m,d_i\right), d_i \in \mathcal{D}
    \label{eq:DAG}
\end{equation}

\begin{equation}
  DAR\left(d_i\right) = \frac{1}{|\mathcal{M}|}\sum_{m \in \mathcal{M}}{\mathbf{1}\left(accuracy_{JPAF}\left(m,d_i\right) = accuracy_{baseline}\left(m,d_i\right)\right)}, d_i \in \mathcal{D}
  \label{eq:DAR}
\end{equation}
Here, $\mathbf{1}(\cdot)$ is the indicator function which equals 1 if the condition is true and 0 otherwise. The terms $accuracy_{JPAF}(m,d_i)$ and $accuracy_{baseline}(m,d_i)$ denote the dimension accuracy of JPAF and the baseline model, respectively, for MBTI type $m$ and dimension $d_i$.

\subsubsection{Activation and Evolution Metrics (Experiment 2)}  
Experiment 2 evaluated JPAF in type-specific challenge scenarios designed to trigger personality adaptations. Evaluation focused on two aspects: whether the intended type was correctly activated and whether subsequent personality shifts followed theoretically valid patterns. Personality shifts were categorized as dominant replacement, auxiliary replacement, dominant–auxiliary swap, structural reorganization, or no change, as defined in Section \ref{sec:reflection}. Two evaluation metrics were used: 

\begin{itemize}
    \item Type Activation Accuracy (TAA): Proportion of cases where a challenge scenario successfully activated the intended target type through the reinforcement–compensation mechanism. High TAA indicates both effective scenario design and correct short-term adaptation.  
    \item Personality Shift Accuracy (PSA): Proportion of cases where personality restructuring followed theoretically valid patterns. PSA reflects the effectiveness of the reflection mechanism in managing long-term adaptation.  
\end{itemize}

\section{Results}
\subsection{Experiment 1: Personality Alignment}
We evaluated whether JPAF's dominant–auxiliary coordination mechanism improves personality alignment compared to a baseline across three LLMs (GPT, Llama, and Qwen). Alignment was measured using dimension accuracy on two MBTI questionnaires (MBTI-93 and MBTI-70), averaged over five runs (see Appendix \ref{tab:Accuracy93gpt}-\ref{tab:Accuracy70qwen} for detailed results). DAG and DAR were subsequently computed to quantify the relative improvement of JPAF over the baseline.

\begin{figure}[htbp]
    \centering
    \label{fig:dag-overall}
    \subfigure[MBTI-93]{
        \includegraphics[width=0.45\linewidth]{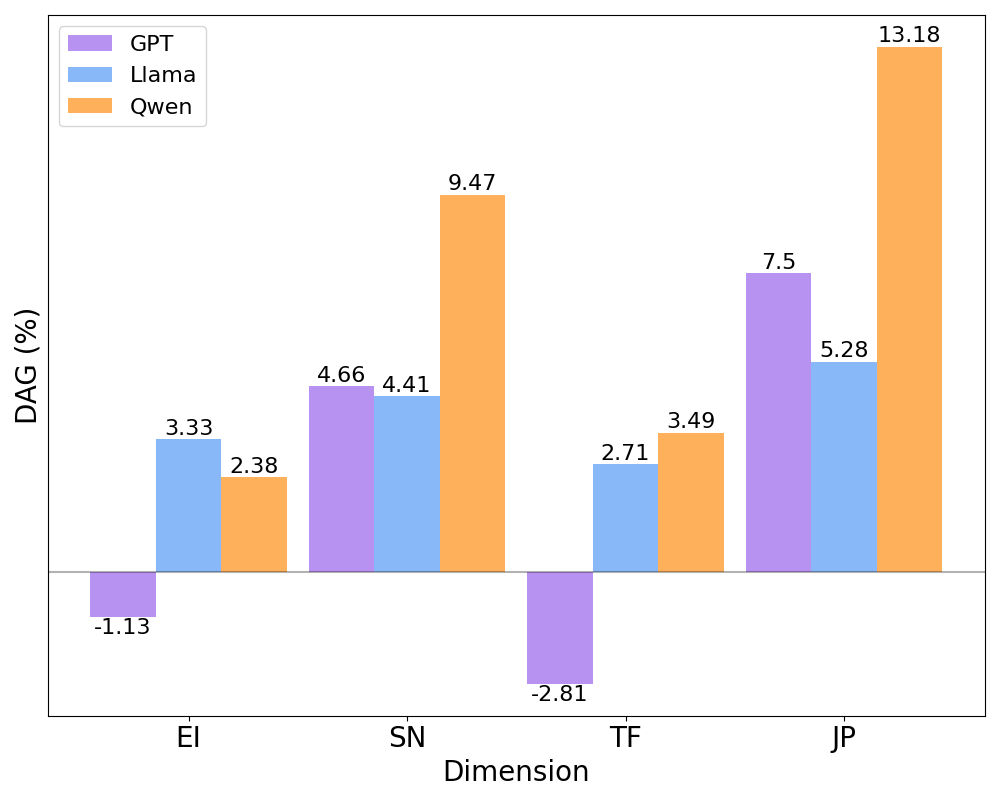}
        \label{fig:dag-93}
    }
    \subfigure[MBTI-70]{
        \includegraphics[width=0.45\linewidth]{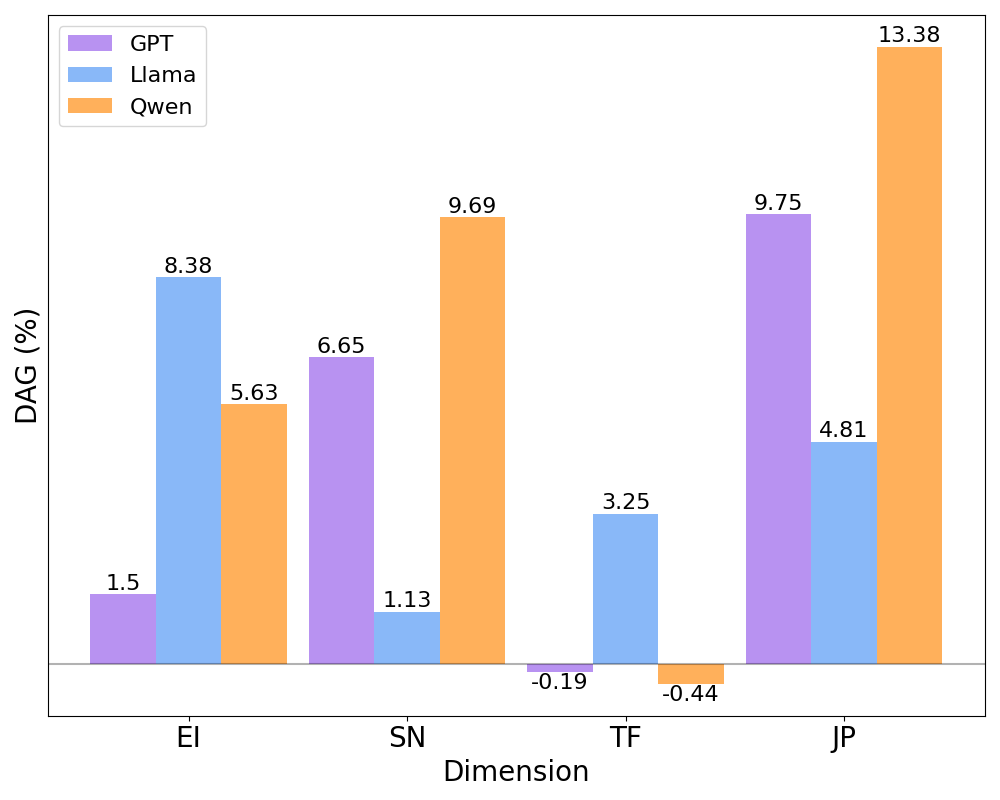}
        \label{fig:dag-70}
    }
    \caption{DAG across models and MBTI dimensions for MBTI-93- and MBTI-70 tests. Positive DAG indicates that JPAF improves personality-aligned responses over the baseline, with the largest gains observed in SN and JP.}
\end{figure}

Figures \ref{fig:dag-93} and \ref{fig:dag-70} show the DAG values for GPT, Llama, and Qwen on the MBTI-93 and MBTI-70 tests, respectively. Positive DAG values indicate improved alignment under JPAF. Across both questionnaires, JPAF generally increased alignment in most MBTI dimensions, with the largest gains observed in JP and SN. For example, on MBTI-70, GPT and Qwen achieved DAG improvements of +9.75\% and +13.38\% in JP, respectively. Since JP links cognitive preferences (IE, SN, TF) to observable behaviors, these results suggest that dominant–auxiliary coordination mechanism is particularly effective for modeling JP traits. In contrast, the TF dimension showed minimal gains, and in some cases the baseline slightly outperformed JPAF (e.g., GPT: –0.19\%, Qwen: –0.44\% on MBTI-70), suggesting that models already handle TF-related judgments relatively well when MBTI type is explicitly prompted.

Model-specific trends emerged as well. GPT showed occasional declines in EI and TF on MBTI-93 (–1.13\% and –2.81\% DAG, respectively), suggesting stronger baseline reasoning in these dimensions. However, GPT with JPAF showed consistent improvements in JP and SN. Llama demonstrated positive but less stable effects of JPAF: while positive DAG values were observed across all dimensions, results varied more strongly between the MBTI-93 and MBTI-70 tests, reflecting less predictable reasoning stability. Qwen benefited most consistently from JPAF, achieving positive DAG across nearly all dimensions, with only a minor decline in TF on MBTI-70, highlighting the robustness of JPAF for this model.

\begin{figure}[htbp]
    \centering
    \subfigure[MBTI-93]{
        \includegraphics[width=0.45\linewidth]{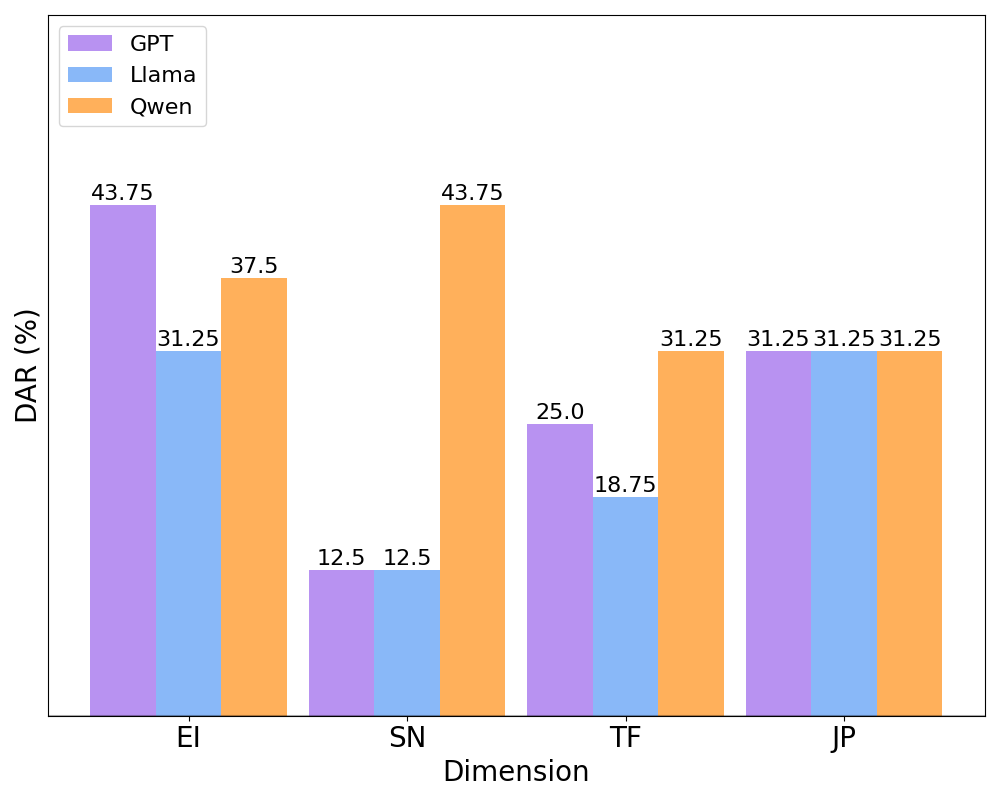}
        \label{fig:dar-93}
    }
    \subfigure[MBTI-70]{
        \includegraphics[width=0.45\linewidth]{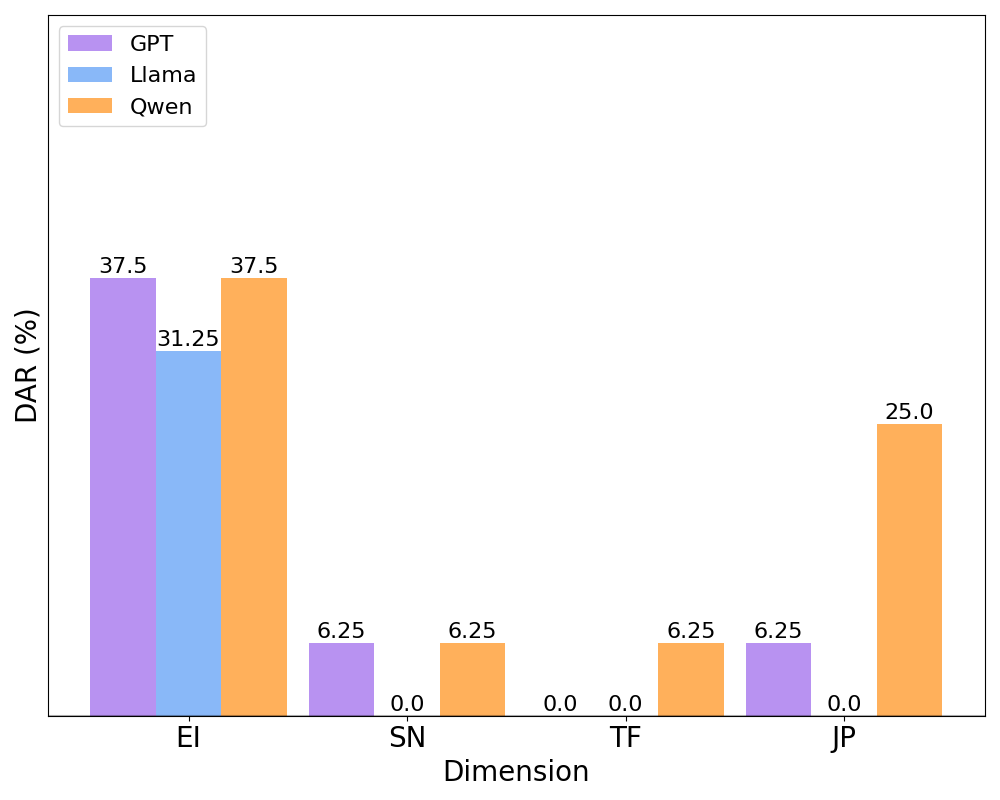}
        \label{fig:dar-70}
    }
    \caption{DAR across models and dimensions for MBTI-93- and MBTI-70 tests. High DAR in EI and JP indicates consistent agreement between JPAF-enhanced and baseline predictions, with JPAF particularly stabilizing SN and JP performance.}

\end{figure}

DAR, shown in Figures \ref{fig:dar-93} and \ref{fig:dar-70}, quantifies agreement between JPAF-enhanced and baseline predictions at the dimension level. High DAR indicates that both models consistently reach the same accuracy for a given dimension, with many cases arising when both achieve 100\% accuracy. On MBTI-93, GPT, Llama, and Qwen all exhibited high DAR in all dimensions, suggesting that JPAF can accurately capture MBTI traits even without explicit type label in prompts. Qwen showed stable DAR across all four dimensions (>30\%), while Llama and GPT achieved such high agreement only in EI and JP. Both Llama and GPT showed lower DAR in SN and TF (12.5–25\%), reflecting greater variability in these dimensions. On MBTI-70, all models maintained DAR above 30\% in EI, consistent with MBTI-93 results and indicating robustness in this dimension. However, Llama's DAR dropped to 0 in all other dimensions, reflecting instability across test formats. GPT and Qwen showed moderate improvements in these dimensions, with Qwen reaching 25\% in JP. Overall, these results indicate that JPAF can improve and, in some cases, stabilize performance across dimensions, though model-specific variability remains.

\begin{figure}[htbp]
    \centering
    \includegraphics[width=0.7\linewidth]{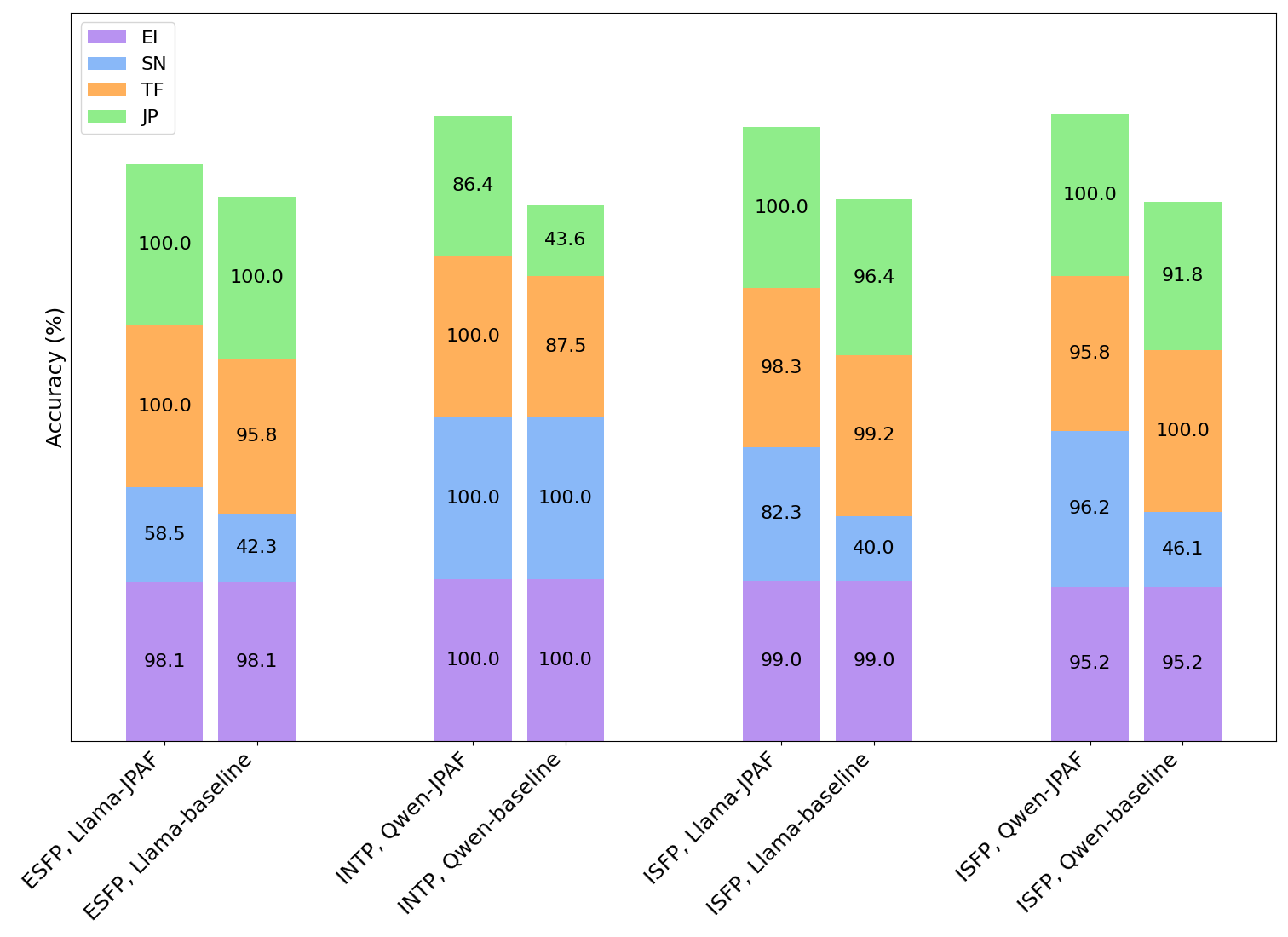}
    \caption{Case-level errors on MBTI-93. Each bar shows four dimension accuracies stacked together, where JPAF models achieved accuracies above 50\% across all dimensions, while baseline models had the SN or JP dimension below 50\%.}
    \label{fig:case93}
\end{figure}

\begin{figure}[htbp]
    \centering
    \includegraphics[width=0.7\linewidth]{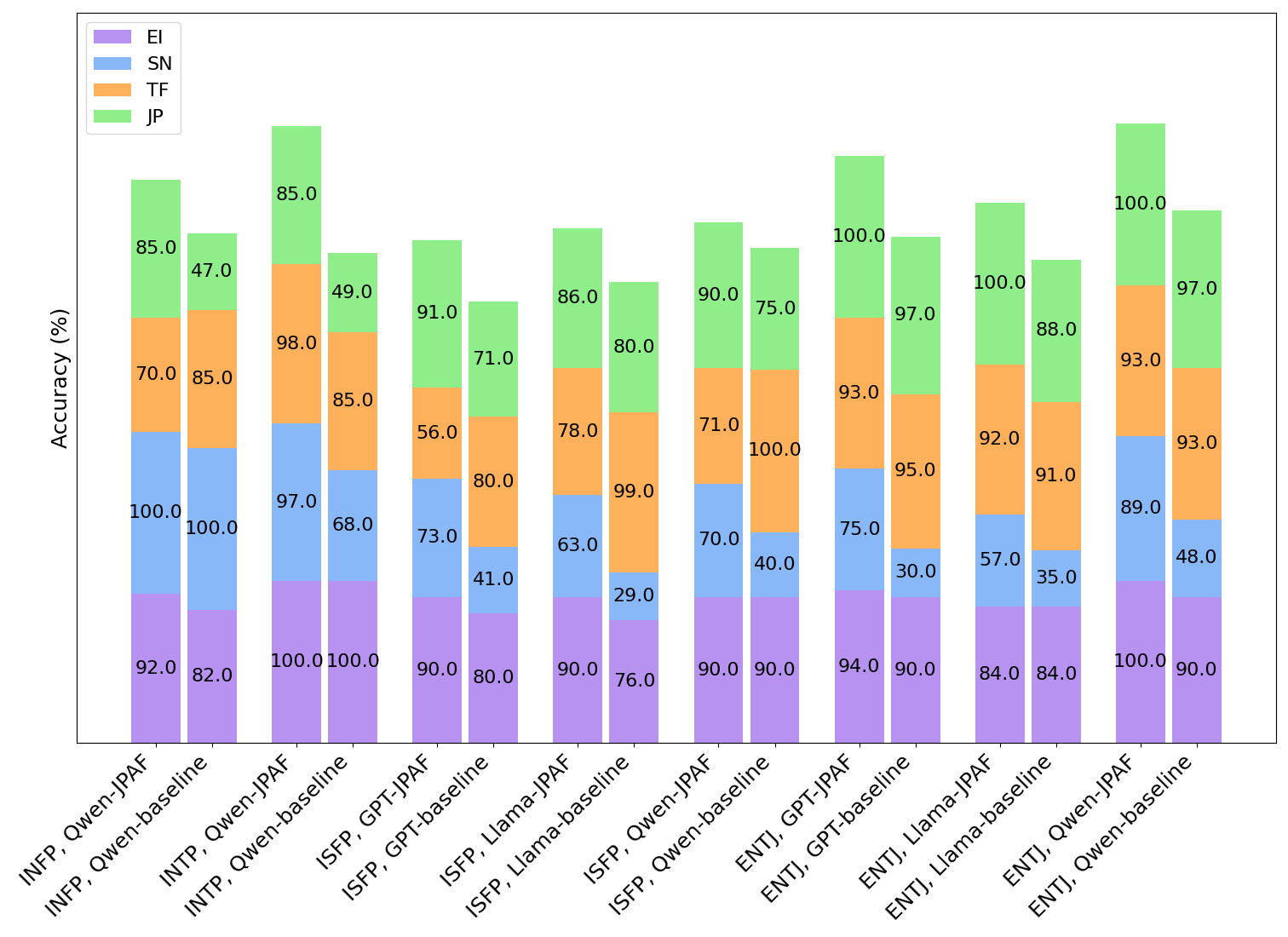}
    \caption{Case-level errors on MBTI-70. Errors are concentrated in SN and JP dimensions, while JPAF improves dimension accuracies.}
    \label{fig:case70}
\end{figure}

Case-level analyses further illustrate JPAF's impact. Without JPAF, models sometimes fell below 50\% accuracy in SN or JP dimensions, even when prompted with the correct MBTI type. Figure \ref{fig:case93} shows cases where Llama and Qwen baseline methods did not reach full alignment across all dimension (<50\%) on MBTI-93. In these cases, JPAF raised all dimension accuracies above 50\%, restoring accurate personality alignment. For example, Llama achieved 42.31\% on SN for ESFP, increasing to 58.5\% with JPAF. For ISFP, Llama and Qwen fell below 50\% on SN at baseline, which improved to 82.31\% and 96.15\% with JPAF, achieving accurate personality alignment. Figure \ref{fig:case70} shows the error cases for GPT, Llama, and Qwen on MBTI-70. At baseline, all models fell below 50\% dimension accuracies in SN for ISFP and ENTJ. Applying JPAF increased SN accuracies, restoring alignment for these personality types. Qwen baseline also fell below 50\% in JP for INFP and INTP, which was corrected after JPAF. 

Overall, Experiment 1 demonstrates that JPAF's dominant-auxiliary coordination mechanism enhances personality alignment across LLMs and across MBTI-93 and MBTI-70 tests, particularly in the SN and JP dimensions, as reflected by reduced errors in MBTI questionnaire responses. Notably, all models using JPAF achieved 100\% accuracy in MBTI personality alignment.

\subsection{Experiment 2: Adaptive and Evolving Personality}
Experiment 2 evaluates how JPAF adapted personality expression in type-specific challenge scenarios through reinforcement–compensation and reflection mechanisms.
We continue to select GPT, Llama, and Qwen as the test models, and apply the eight scenario-based test questions to each model with JPAF enabled, primarily to examine the application of the reinforcement-compensation within JPAF. Tables \ref{tab:select-gpt}-\ref{tab:select-qwen} present the TAA for the three models across different scenarios after incorporating JPAF. Overall, GPT again performed best, achieving over 90\% accuracy across all eight scenarios. Notably, GPT reached 100\% accuracy in the Fi and Se scenarios, and exceeded 99\% in both the Fe and Te scenarios. Qwen ranked second, with accuracy above 90\% in seven scenarios, including 100\% in Se and Ne, and over 99\% in Fe. However, its performance dropped to 88.75\% in the Ti scenario. Llama showed the weakest performance, with TAA exceeding 90\% only in Fi and Se, while remaining between 65\% and 80\% across the other six scenarios, particularly struggling in Ti and Te. Analysis of the outputs reveals that Llama was unable to distinguish Ti and Te as precisely as GPT and Qwen, resulting in incorrect type selections. Similar issues also occurred in the Ni and Ne scenarios. Despite these inconsistencies in Llama (65\%–80\%), the consistently high TAA of GPT and Qwen (mostly above 90\%) demonstrates that the eight scenario-based test cases effectively activate the reinforcement-compensation mechanism within JPAF.

\begin{figure}[htbp]
    \centering
    \includegraphics[width=0.7\linewidth]{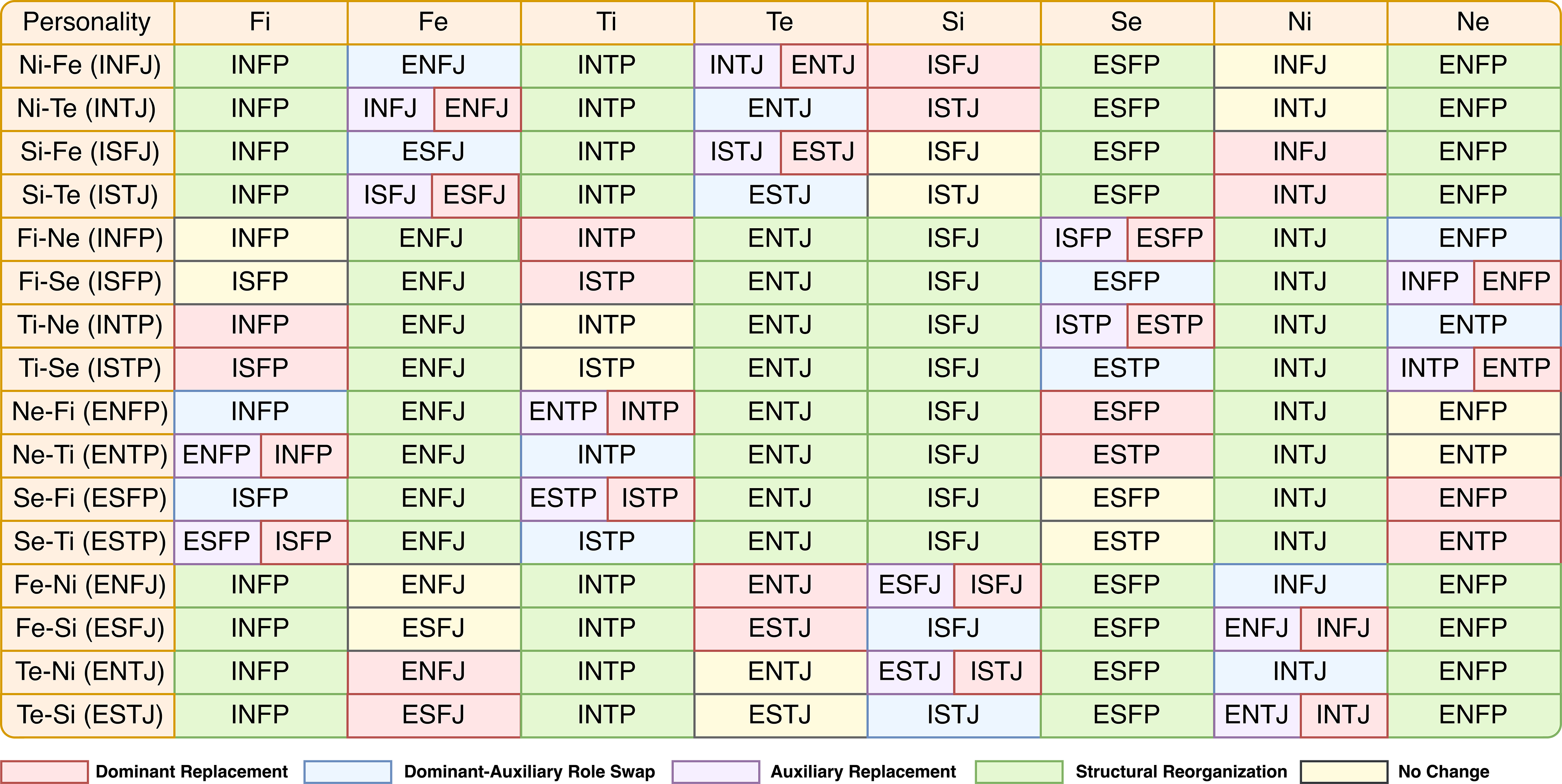}
    \caption{Personality changes in 8 scenarios based on the GPT model}
    \label{fig:gpt_change}
\end{figure}

\begin{figure}[htbp]
    \centering
    \includegraphics[width=0.7\linewidth]{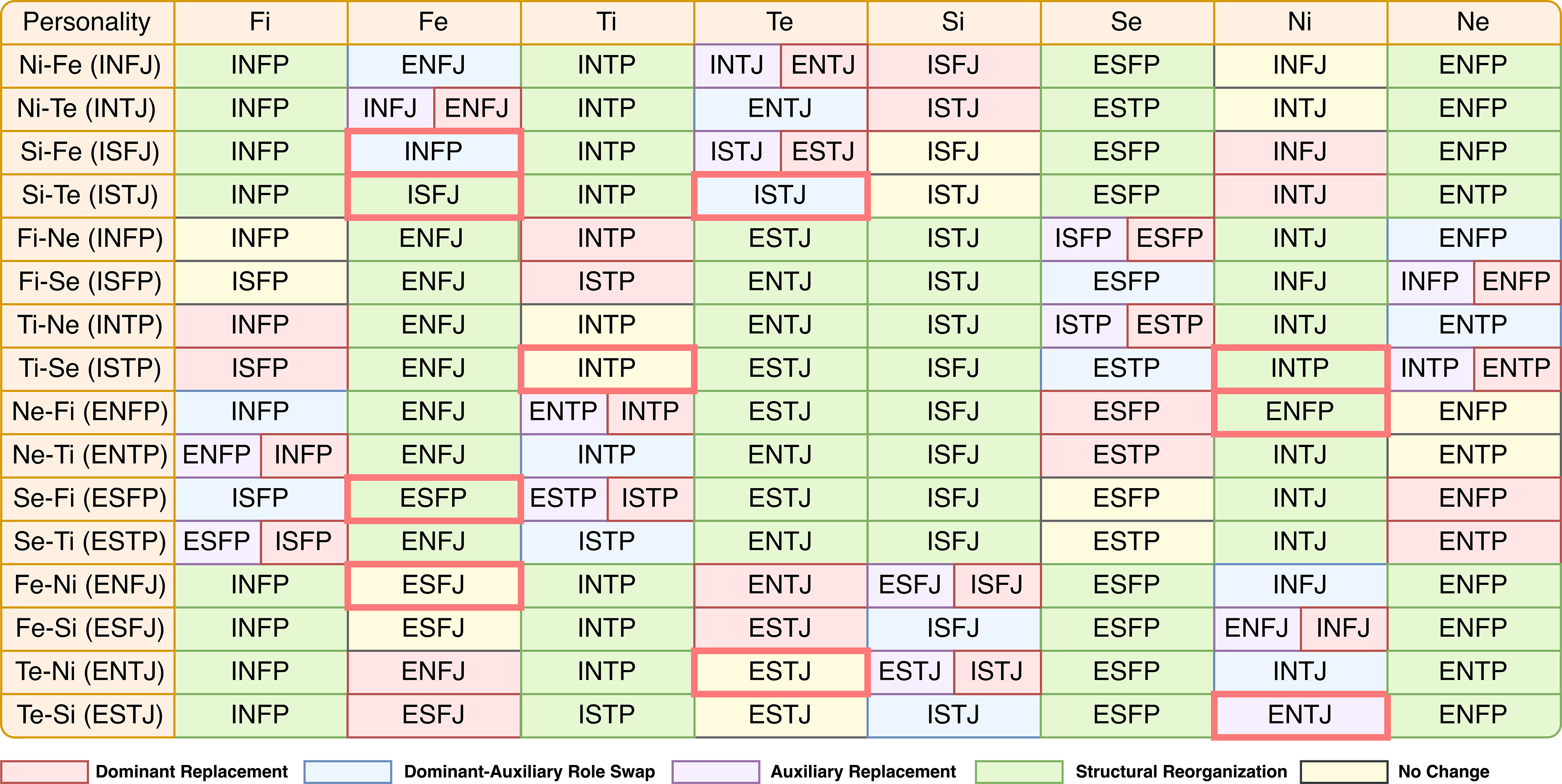}
    \caption{Personality changes in 8 scenarios based on the Llama model}
    \label{fig:llama_change}
\end{figure}

\begin{figure}[htbp]
    \centering
    \includegraphics[width=0.7\linewidth]{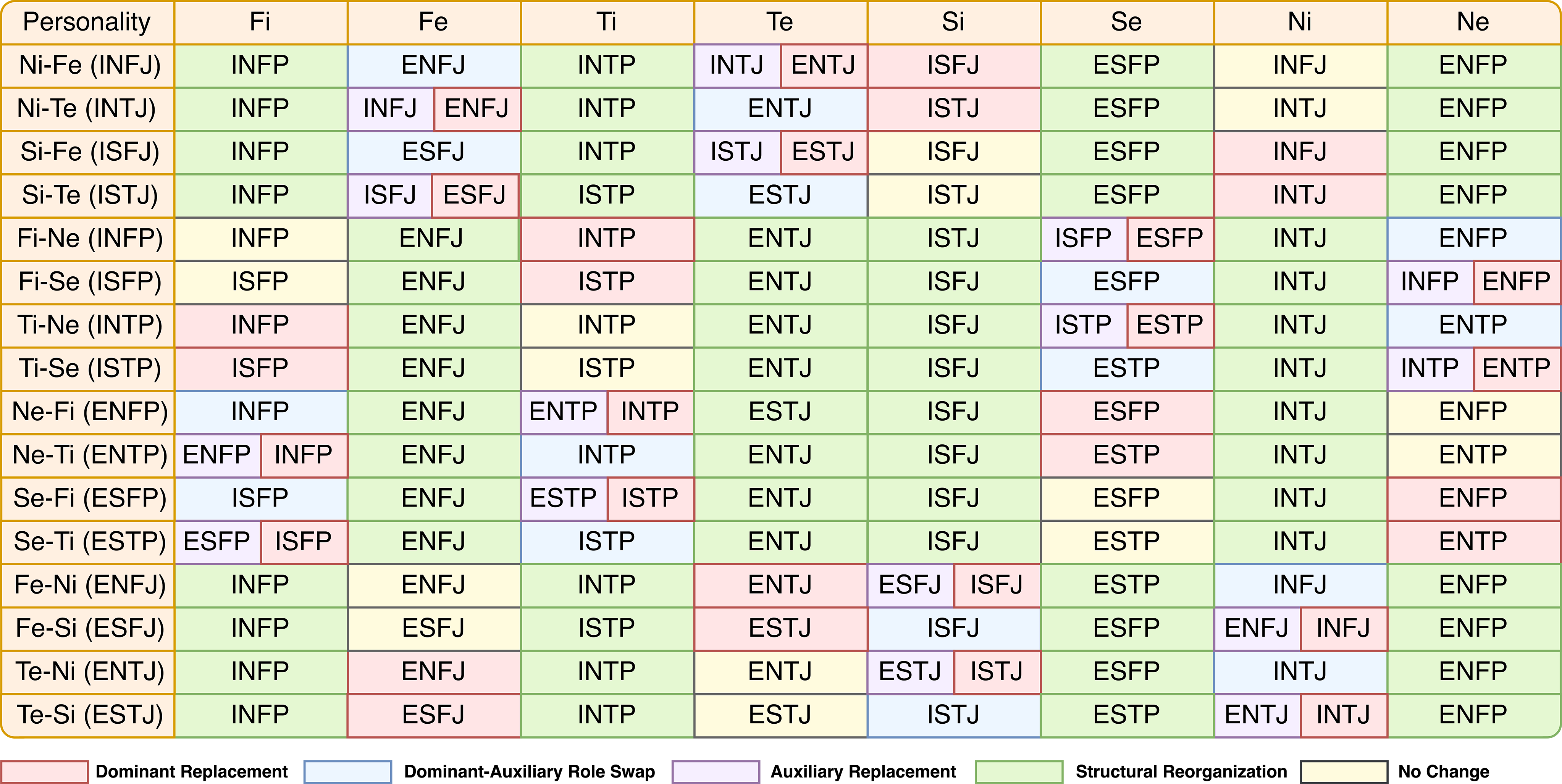}
    \caption{Personality changes in 8 scenarios based on the Qwen model}
    \label{fig:qwen_change}
\end{figure}

Figures \ref{fig:gpt_change}-\ref{fig:qwen_change}, illustrate the PSA of the three models after applying JPAF to the eight scenarios. As defined in Sections \ref{sec:reinforcement} and \ref{sec:reflection}, when the reinforcement–compensation mechanism is triggered, reflection mechanism  is likely to be activated, potentially leading to personality changes. To make these changes more visible, we use five colors in the figures to represent Dominant Replacement, Auxiliary Replacement, Dominant–Auxiliary Role Swap, Structural Reorganization, and No Change. Across Figures \ref{fig:gpt_change}-\ref{fig:qwen_change}, it can be observed that GPT, Llama, and Qwen underwent personality changes under different scenarios, with certain patterns emerging. GPT and Qwen both achieved 100\% PSA, showing complete consistency in Dominant Replacement, Auxiliary Replacement, Dominant–Auxiliary Role Swap, and No Change. However, they differed slightly in Structural Reorganization. For instance, GPT's Si-Te (ISTJ) shifted to Ti-Ne (INTP) in the Ti scenario, while Qwen's Si-Te (ISTJ) shifted to Ti-Se (ISTP). Both Ti-Ne (INTP) and Ti-Se (ISTP) were correct personality changes, but the divergence arose from the auxiliary tpye each model selected for Ti, based on its historical data. Llama, by contrast, exhibited a higher rate of errors. Ten personality changes were incorrect, highlighted with red boxes in Figure \ref{fig:llama_change}. The primarily reason was that Llama was unable to reliably use JPAF to select the correct psychological type under different scenarios. Taking Si-Te (ISTJ) as an example, its TAA in the Fe scenario was only 33\% (see Table \ref{tab:select-llama}). As a result, after responding to 15 Fe-related questions, Si-Te (ISTJ) was incorrectly transformed into Si-Fe (ISFJ). Similar errors occurred in other 9 cases, which attributed to Llama's lower TAA, leading to incorrect personality transformations. Nevertheless, Llama still maintains an overall accuracy rate of 92.19\% in PSA.


\begin{figure}[htbp]
    \centering
    \includegraphics[width=0.7\linewidth]{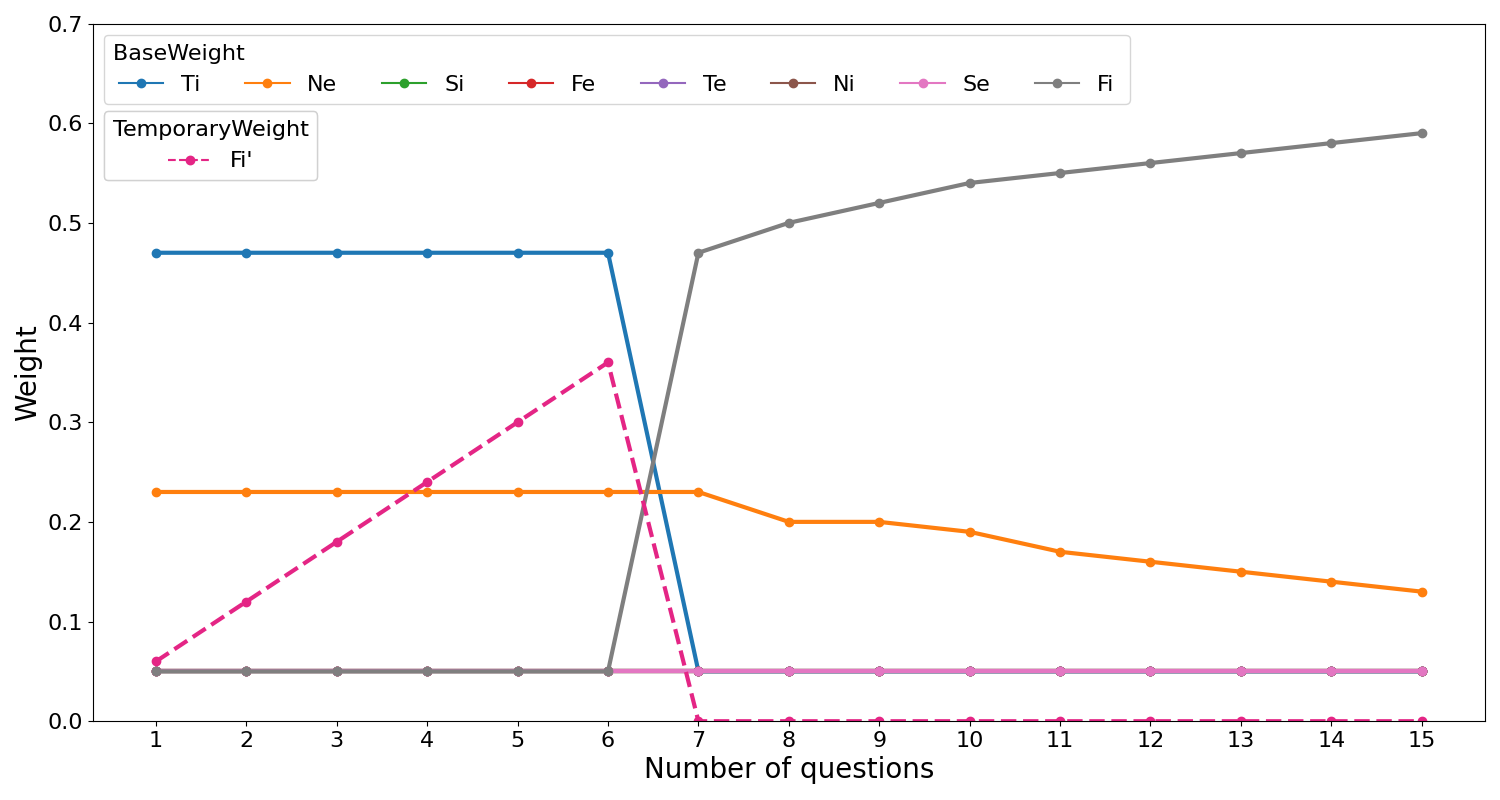}
    \caption{The weight change of Ti-Ne (INTP) personality in the Fi scenario}
    \label{fig:intp-fi}
\end{figure}

\begin{figure}[htbp]
    \centering
    \includegraphics[width=0.7\linewidth]{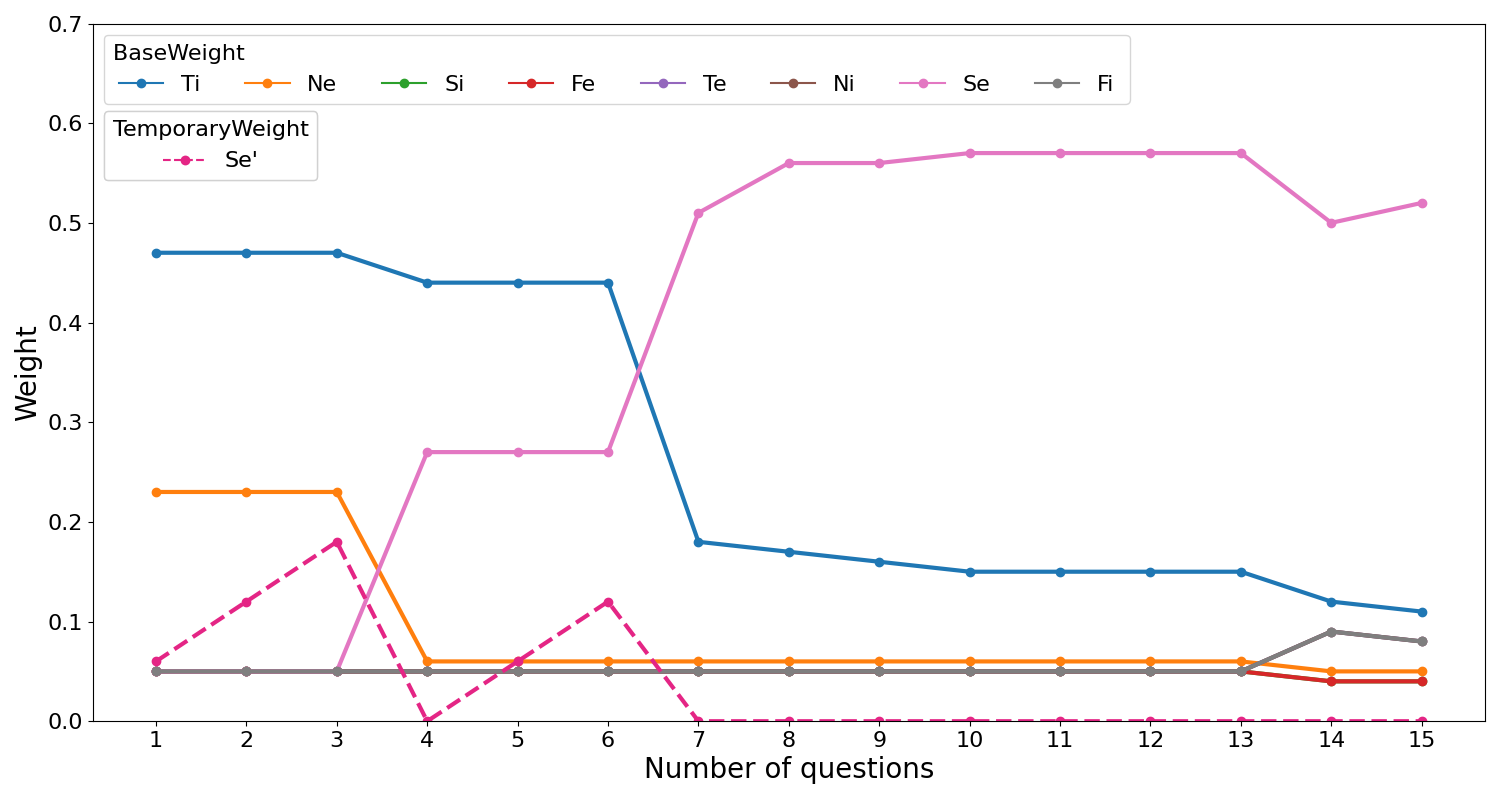}
    \caption{The weight change of Ti-Ne (INTP) personality in the Se scenario}
    \label{fig:intp-se}
\end{figure}

\begin{figure}[htbp]
    \centering
    \includegraphics[width=0.7\linewidth]{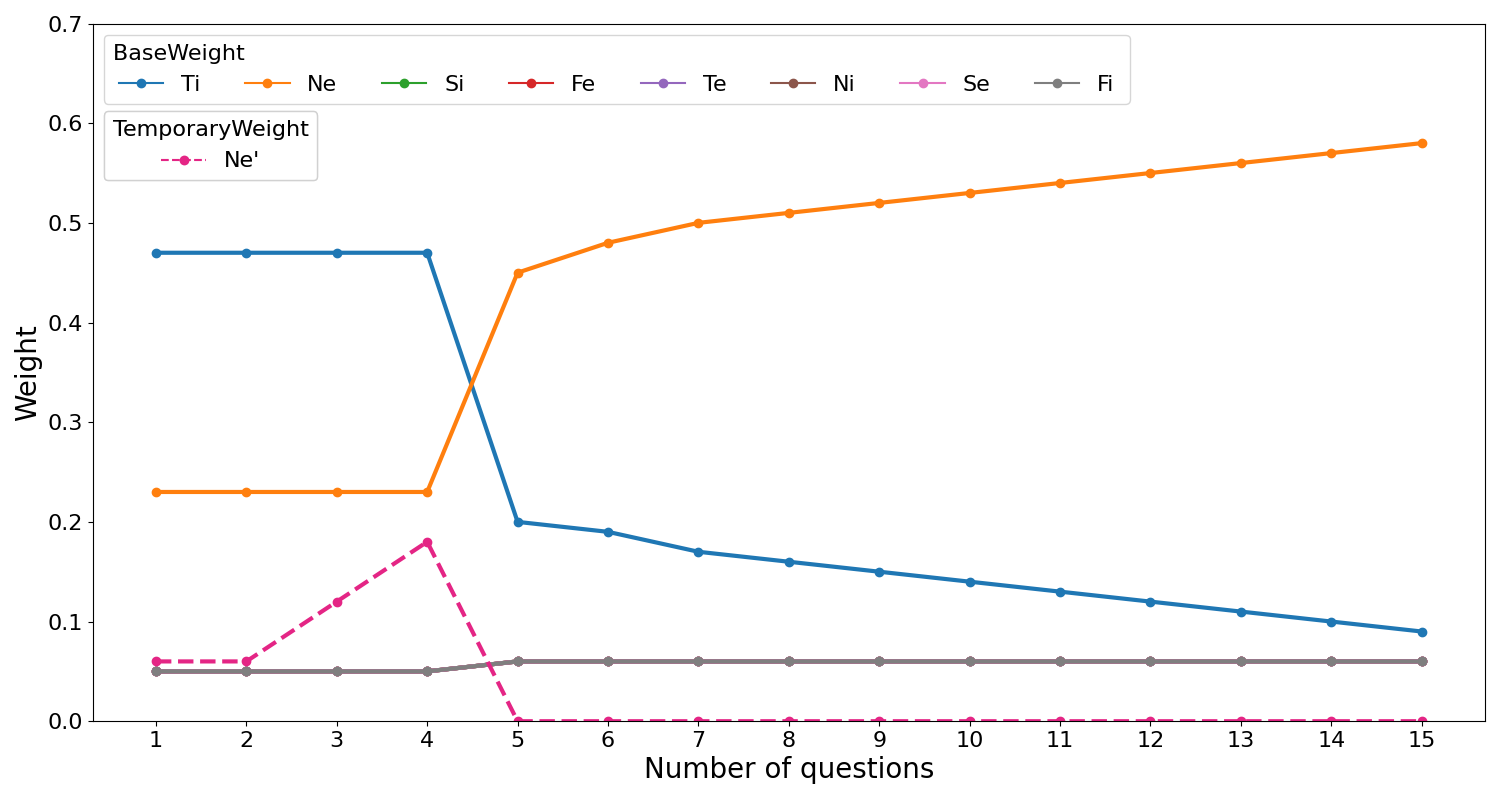}
    \caption{The weight change of Ti-Ne (INTP) personality in the Ne scenario}
    \label{fig:intp-ne}
\end{figure}

\begin{figure}[htbp]
    \centering
    \includegraphics[width=0.7\linewidth]{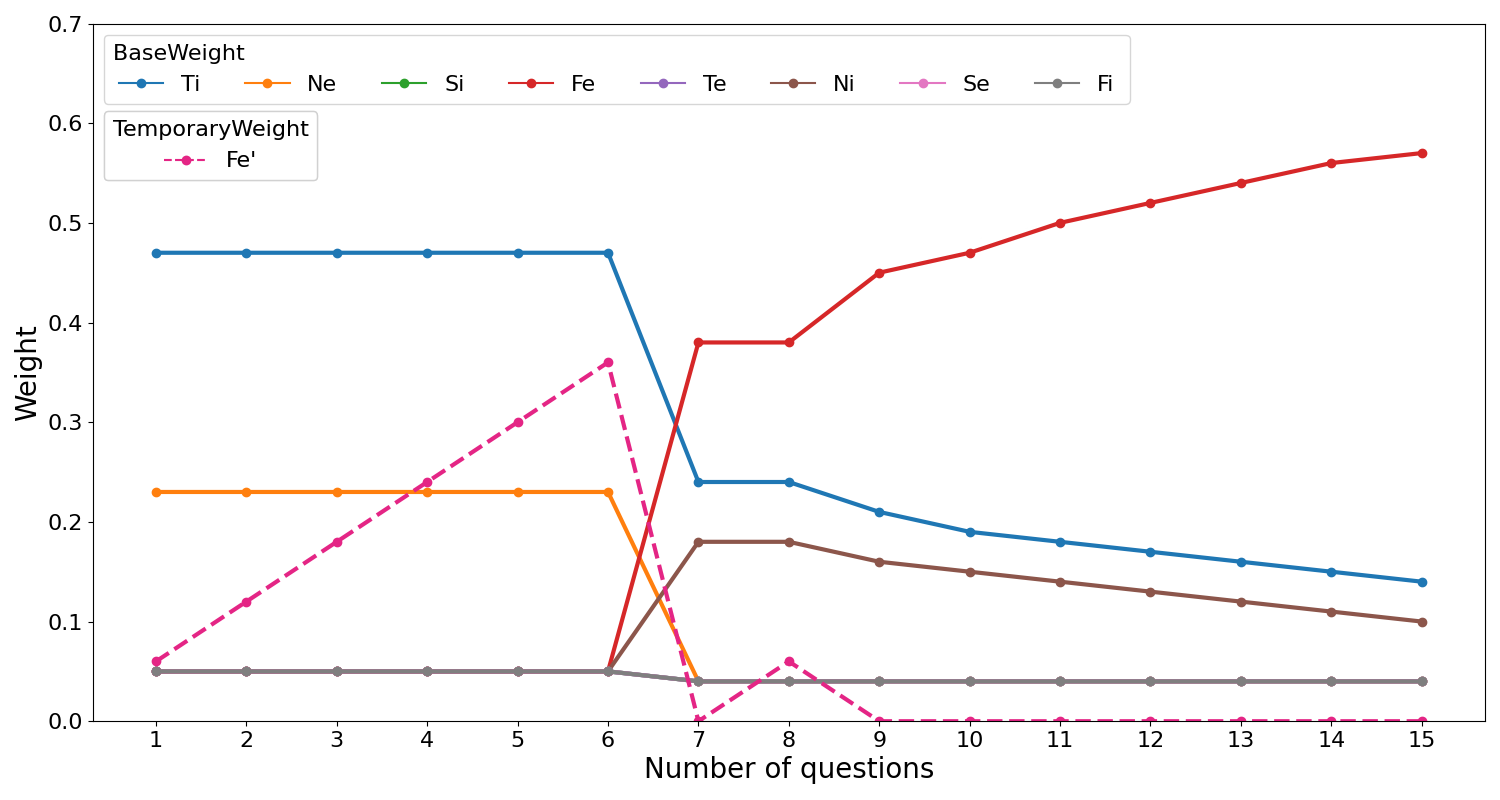}
    \caption{The weight change of Ti-Ne (INTP) personality in the Fe scenario}
    \label{fig:intp-fe}
\end{figure}

\begin{figure}[htbp]
    \centering
    \includegraphics[width=0.7\linewidth]{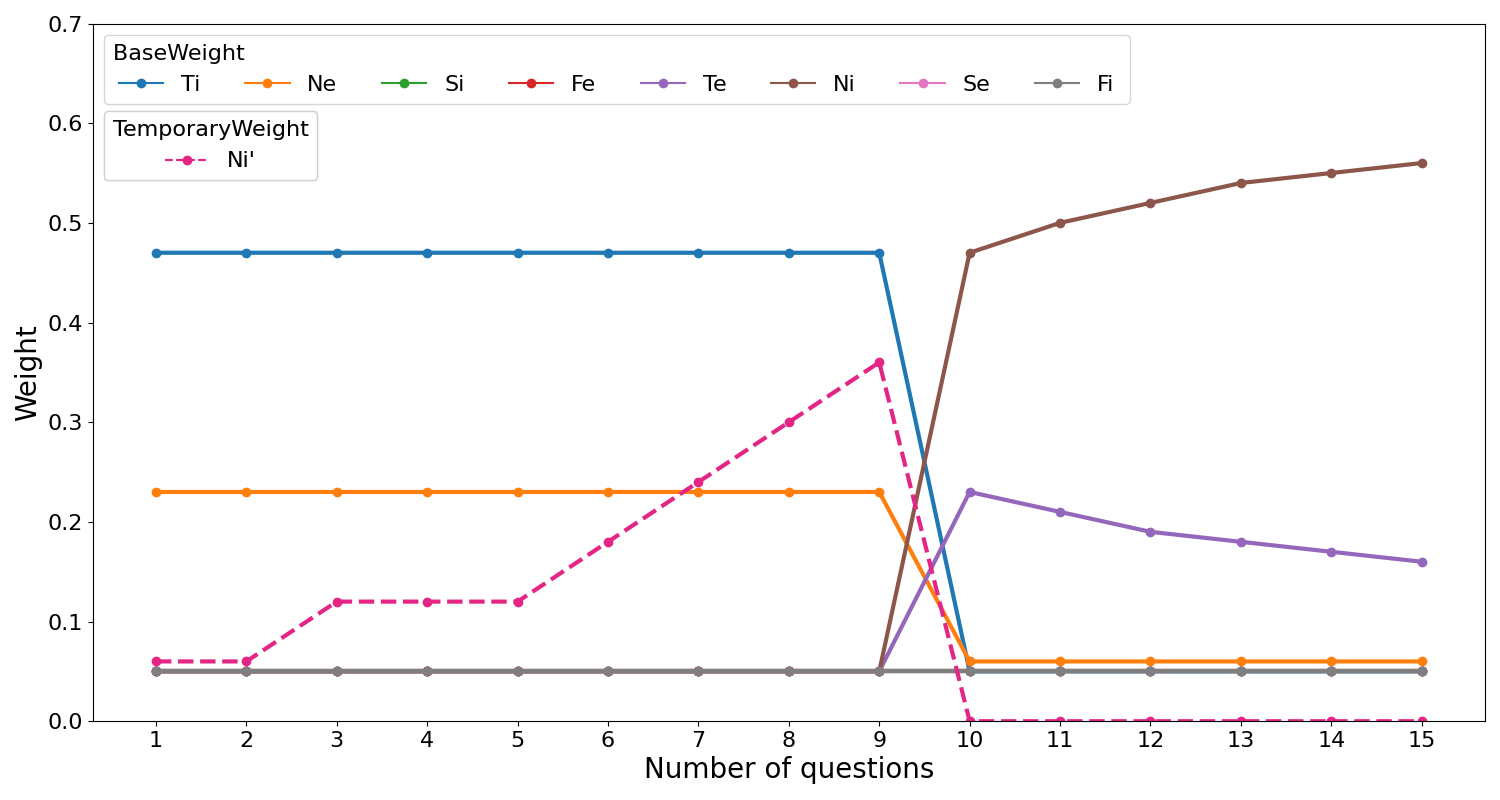}
    \caption{The weight change of Ti-Ne (INTP) personality in the Ni scenario}
    \label{fig:intp-ni}
\end{figure}

\begin{figure}[htbp]
    \centering
    \includegraphics[width=0.7\linewidth]{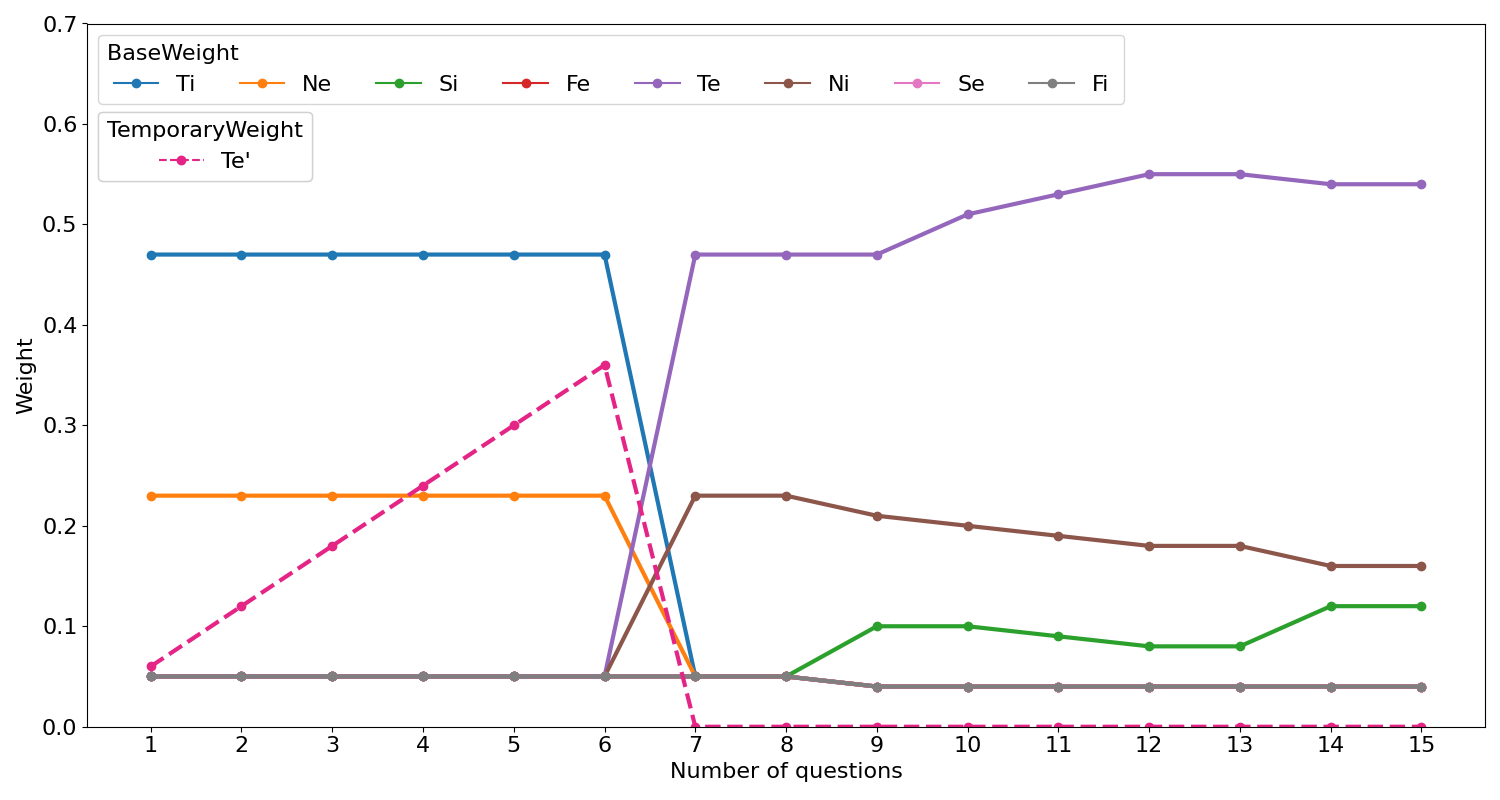}
    \caption{The weight change of Ti-Ne (INTP) personality in the Te scenario}
    \label{fig:intp-te}
\end{figure}

\begin{figure}[htbp]
    \centering
    \includegraphics[width=0.7\linewidth]{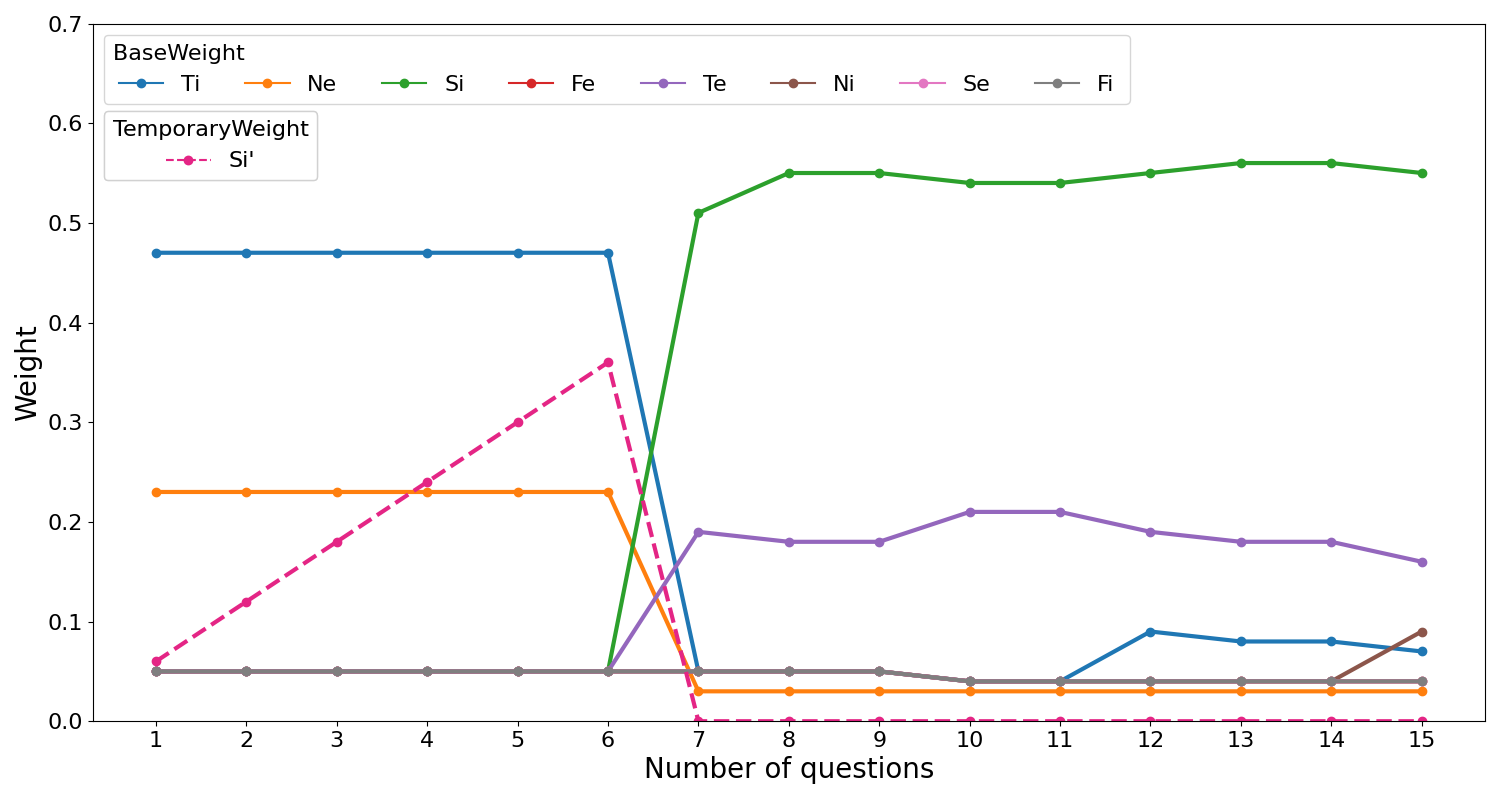}
    \caption{The weight change of Ti-Ne (INTP) personality in the Si scenario}
    \label{fig:intp-si}
\end{figure}

\begin{figure}[htbp]
    \centering
    \includegraphics[width=0.7\linewidth]{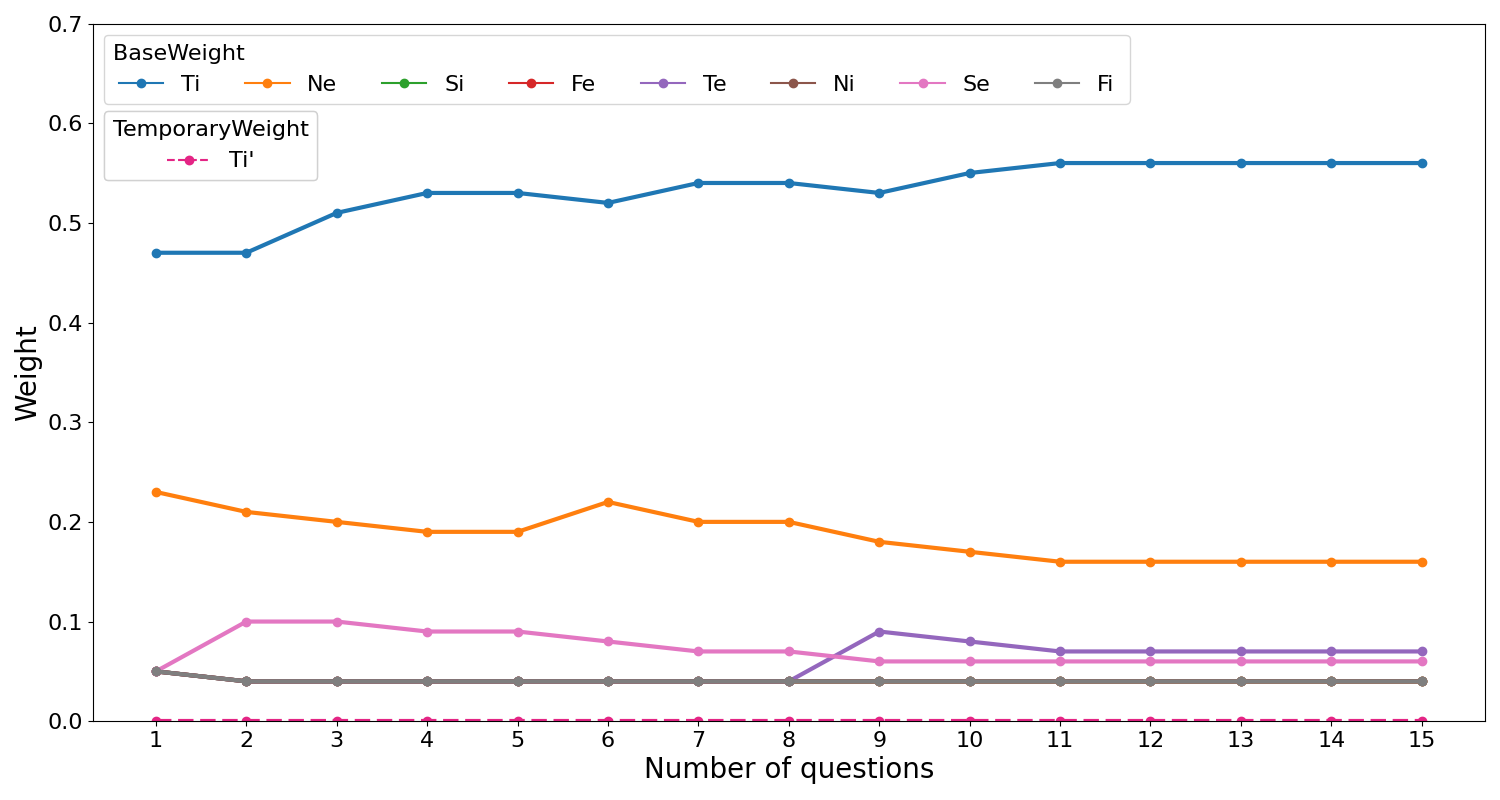}
    \caption{The weight change of Ti-Ne (INTP) personality in the Ti scenario}
    \label{fig:intp-ti}
\end{figure}

Below, we take Ti-Ne (INTP) from the GPT data as an example to illustrate the transformation process in detail. Figures \ref{fig:intp-fi}-\ref{fig:intp-ti} present the changes in type weights during the evolution of Ti-Ne (INTP).

When Ti-Ne (INTP) encounters an Fi scenario (Fi and dominant type Ti belong to the same attitude but represent different types), the reinforcement–compensation mechanism continuously activates the undifferentiated Fi of Ti-Ne (INTP). As a result, the \emph{TemporaryWeight} of Fi increases steadily, reflection mechanism is triggered, and eventually a Dominant Replacement occurs. As shown in Figure \ref{fig:intp-fi}, the \emph{TemporaryWeight} of Fi raised progressively, and by the completion of the seventh question, the sum of Fi's \emph{TemporaryWeight} and \emph{BaseWeight} surpassed the \emph{BaseWeight} of Ti. Fi thus replaced Ti as the dominant type, with Fi's \emph{BaseWeight} increasing while Ti's \emph{BaseWeight} decreases. This reconfiguration formed a new dominant–auxiliary structure of Fi-Ne, transforming Ti-Ne (INTP) into Fi-Ne (INFP). Since Fi had became the dominant function, once its \emph{BaseWeight} reached or exceeded 0.5, the normalization mechanism within reflection mechanism (see Algorithm \ref{al:jpaf}: JPAF Algorithm) was repeatedly triggered, resetting Fi's \emph{TemporaryWeight} to zero. Consequently, as shown in Figure \ref{fig:intp-fi}, Fi's \emph{TemporaryWeight} was reset to zero at the seventh question and no longer changes thereafter (the same applies to subsequent cases, and the \emph{TemporaryWeight} resetting will not be further elaborated).

When Ti-Ne (INTP) encounters an Se scenario (Se and auxiliary type Ne belong to different types under the same attitude), the reinforcement-compensation mechanism continuously activates the undifferentiated Se in Ti-Ne (INTP). As a result, the \emph{TemporaryWeight} of Se gradually increases, triggering reflection mechanism and ultimately leading to two outcomes: Auxiliary Replacement and Dominant Replacement. As shown in Figure \ref{fig:intp-se}, after answering the fourth question, the combined value of Se's temporary and \emph{BaseWeight} exceeded Ne's \emph{BaseWeight}. Reflection mechanism was triggered, and Se replaced Ne as the new auxiliary function. Consequently, Se's \emph{BaseWeight} increased while Ne's \emph{BaseWeight} decreased, forming a new dominant–auxiliary structure of Ti-Se. At this stage, Ti-Ne (INTP) shifted to Ti-Se (ISTP), marking the Auxiliary Replacement. However, after answering the seventh question, due to the continued reinforcement of Se, the sum of its base and \emph{TemporaryWeight}s surpassed the \emph{BaseWeight} of the dominant Ti. Reflection mechanism was triggered again, and Se replaced Ti as the new dominant type, while Ti shifted to auxiliary. This formed a new dominant–auxiliary structure of Se-Ti, resulting in a transition from Ti-Se (ISTP) to Se-Ti (ESTP). Throughout the process, under the persistent activation of the Se scenario, the structure of Ti-Ne (INTP) first shifted to Ti-Se (ISTP) and finally stabilized at Se-Ti (ESTP), with two personality transformations occurring in sequence.

When Ti-Ne (INTP) encounters an Ne scenario, since Ne is the auxiliary type, the reinforcement-compensation mechanism continuously strengthens the \emph{TemporaryWeight} of Ne. Once reflection mechanism is triggered, a Dominant–Auxiliary Role Swap occurs. As shown in Figure \ref{fig:intp-ne}, after completing the fifth question, the combined value of Ne's \emph{TemporaryWeight} and \emph{BaseWeight} surpassed the \emph{BaseWeight} of the dominant function Ti. Reflection mechanism was then activated, leading Ne to become the dominant type while Ti shifts into the auxiliary type. Consequently, Ne's \emph{BaseWeight} increased while Ti's \emph{BaseWeight} decreased, forming a new dominant–auxiliary structure of Ne-Ti, thereby transforming Ti-Ne (INTP) into Ne-Ti (ENTP).

Figures \ref{fig:intp-fe}-\ref{fig:intp-si} illustrate cases of Structural Reorganization occurring in Ti-Ne (INTP). Taking the Fe scenario as an example, when Ti-Ne (INTP) encounters Fe scenario, the reinforcement–compensation mechanism repeatedly activates Fe to compensate for the scenario. Once reflection mechanism is triggered, since Fe cannot form a stable psychological type structure with Ti-Ne, Structural Reorganization occurs. As shown in Figure \ref{fig:intp-fe}, after responding to the seventh question, the combined \emph{TemporaryWeight} and \emph{BaseWeight} of Fe surpassed the \emph{BaseWeight} of Ti. Reflection mechanism was then activated, and Fe was first identified as the dominant type. However, according to Jungian theory in Section \ref{sec:Dominant-Auxiliary}, neither Ti nor Ne could serve as the auxiliary type for Fe. The model therefore assigned an auxiliary type to Fe based on prior data. In this case, Ni was selected as the auxiliary, transforming Ti-Ne (INTP) into Fe-Ni (ENFJ). The \emph{BaseWeight} of both Ti and Ne were subsequently reduced. Although the reduced \emph{BaseWeight} of Ti still remained higher than that of Ni, it posed no threat to Fe's dominance and thus did not disrupt the stable structure of the new Fe-Ni (ENFJ) configuration. Similarly, when Ti-Ne (INTP) encounters Ni, Te, or Si scenarios, Structural Reorganization also occurs. As shown in Figure \ref{fig:intp-ni}, after responding to the tenth question, the combined \emph{TemporaryWeight} and \emph{BaseWeight} of Ni surpassed the \emph{BaseWeight} of Ti, triggering reflection mechanism. Ni was then identified as the dominant function, and the model assigned an auxiliary type based on prior data. In this case, Te was selected, transforming Ti-Ne (INTP) into Ni-Te (INTJ). Under Te and Si scenarios, the process is analogous to that observed in the Fe and Ni cases, with Ti-Ne (INTP) reorganizing into Te-Ni (ENTJ) and Si-Fe (ISFJ), respectively, as illustrated in Figures \ref{fig:intp-te}-\ref{fig:intp-si}.


A special case occurs when Ti-Ne (INTP) encounters the scenario targeting its dominant type Ti, in which the Ti-Ne (INTP) structure remains unchanged. As shown in Figure \ref{fig:intp-ti}, repeated activation of Ti rapidly drove its \emph{BaseWeight} above 0.5, triggering normalization in the reflection mechanism (see Algorithm \ref{al:jpaf}: JPAF Algorithm). In this process, the \emph{BaseWeight} of Ti continued to increase while its \emph{TemporaryWeight} was repeatedly reset to zero. Meanwhile, the \emph{BaseWeight} of the auxiliary type Ne decreased correspondingly, but the overall Ti-Ne (INTP) structure remained unaffected.

\section{Conclusion}
This paper introduced the JPAF, a psychologically grounded approach for modeling, adapting, and evolving LLM personalities. By integrating dominant–auxiliary coordination, reinforcement–compensation, and reflection mechanisms, JPAF achieves near-perfect MBTI alignment while supporting dynamic personality shifts and long-term structural evolution. 
Empirical evaluation across three LLM families showed 100\% MBTI alignment for all models, over 90\% type activation accuracy for GPT and Qwen (and 65–95\% for LLaMA), and theoretically valid personality shifts with 100\% accuracy for GPT and Qwen and 92\% for LLaMA. These findings confirm JPAF’s effectiveness in producing coherent, adaptive, and evolving personalities.
For HCI, these findings highlight how LLM agents can move beyond static role conditioning toward dynamic personalities that evolve through interaction. Such capabilities open design opportunities in education, healthcare, entertainment, and collaborative systems where long-term engagement and trust depend on consistent yet adaptive personalities. 
Future research should integrate affective and cognitive traits, explore richer social and multi-agent contexts, and examine the ethical implications of deploying evolving personalities in real-world settings. Ultimately, adaptive LLM agents with evolving personality point toward a more naturalistic and human-like agent design, paving the way for deeper, sustained human-AI relationships.



\bibliographystyle{ACM-Reference-Format}
\bibliography{sample-base} 

\appendix
\section{Appendix}
\subsection{Dimension Accuracy Results}

\begin{table}[htbp]
    \centering
    \caption{The results of 16 MBTI personality types based on GPT with MBTI-93}
    \begin{tabular}{ccccccccc}
        \toprule
        \multirow{2}{*}{Personality} & \multicolumn{2}{c}{EI}& \multicolumn{2}{c}{SN} & \multicolumn{2}{c}{TF} & \multicolumn{2}{c}{JP}   \\
             & Baseline & JPAF & Baseline & JPAF & Baseline & JPAF & Baseline & JPAF \\
        \midrule
        Ni-Fe (INFJ) & 100.00\% &100.00\% &\textbf{96.15\%} &84.62\% &95.83\% &95.83\% &87.27\% &\textbf{100.00\%} \\
        Ni-Te (INTJ) & 100.00\% &100.00\% &\textbf{94.62\%} &85.38\% &100.00\% &100.00\% &100.00\% &100.00\% \\
        Si-Fe (ISFJ) & 90.48\% &\textbf{95.24\%} &95.38\% &\textbf{100.00\%} &84.17\% &\textbf{95.83\%} &97.27\% &\textbf{100.00\%} \\
        Si-Te (ISTJ) & 95.24\% &95.24\% &\textbf{100.00\%} &97.69\% &95.00\% &\textbf{100.00\%} &100.00\% &100.00\% \\
        Fi-Ne (INFP) & 97.14\% &\textbf{100.00\%} &100.00\% &100.00\% &\textbf{99.17\%} &93.33\% &70.91\% &\textbf{91.82\%} \\
        Fi-Se (ISFP) & 90.48\% &\textbf{97.14\%} &53.85\% &\textbf{91.54\%} &\textbf{96.67\%} &91.67\% &88.18\% &\textbf{100.00\%} \\
        Ti-Ne (INTP) & 100.00\% &100.00\% &\textbf{100.00\% }&99.23\% &\textbf{84.17\%} &76.67\% &59.09\% &\textbf{90.91\%} \\
        Ti-Se (ISTP) & 99.05\% &99.05\% &86.15\% &\textbf{88.46\%} &\textbf{99.17\%} &79.17\% &88.18\% &\textbf{100.00\%} \\
        Ne-Fi (ENFP) & \textbf{100.00\%} &93.33\% &96.15\% &\textbf{100.00\%} &\textbf{95.83\%} &92.50\% &90.91\% &90.91\% \\
        Ne-Ti (ENTP) & \textbf{100.00\%} &93.33\% &100.00\% &100.00\% &\textbf{91.67\%} &78.33\% &\textbf{94.55\%} &93.64\% \\
        Se-Fi (ESFP) & \textbf{100.00\%} &98.10\% &50.00\% &\textbf{72.31\%} &95.00\% &\textbf{96.67\%} &95.45\% &\textbf{100.00\%} \\
        Se-Ti (ESTP) & 100.00\% &100.00\% &\textbf{71.54\%} &66.15\% &\textbf{92.50\%} &70.83\% &\textbf{98.18\%} &92.73\% \\
        Fe-Ni (ENFJ) & \textbf{100.00\%} &95.24\% & 83.08\% &\textbf{88.46\%} &95.00\% &\textbf{95.83\%} &79.09\% &\textbf{100.00\%} \\
        Fe-Si (ESFJ) & 100.00\% &100.00\% &95.38\% &\textbf{99.23\%} &83.33\% &\textbf{95.83\%} &100.00\% &100.00\% \\
        Te-Ni (ENTJ) & \textbf{95.24\%} &88.57\% &56.92\% &\textbf{84.62\%} &100.00\% &100.00\% &90.91\% &\textbf{100.00\%} \\
        Te-Si (ESTJ) & \textbf{94.29\%} &88.57\% &\textbf{100.00\%} &96.15\% &100.00\% &100.00\% &100.00\% &100.00\% \\
        \bottomrule
    \end{tabular}
    \label{tab:Accuracy93gpt}
\end{table}    

\begin{table}[htbp]
    \centering
    \caption{The results of 16 MBTI personality types based on Llama with MBTI-93}
    \begin{tabular}{ccccccccc}
        \toprule
        \multirow{2}{*}{Personality} & \multicolumn{2}{c}{EI}& \multicolumn{2}{c}{SN} & \multicolumn{2}{c}{TF} & \multicolumn{2}{c}{JP}   \\
             & Baseline & JPAF & Baseline & JPAF & Baseline & JPAF & Baseline & JPAF \\
        \midrule
        Ni-Fe (INFJ) & 99.05\% & 99.05\% &   \textbf{100.00\%} & 92.31\% &   91.67\% & \textbf{100.00\%} &   80.00\% & \textbf{100.00\%} \\
        Ni-Te (INTJ) & 100.00\% & 100.00\% &   \textbf{96.15\%} & 84.62\% &   100.00\% & 100.00\% &   98.18\% & \textbf{100.00\%} \\
        Si-Fe (ISFJ) & 86.67\% & \textbf{100.00\%} &   \textbf{100.00\%} & 96.15\% &   83.33\% & \textbf{100.00\%} &   96.36\% & \textbf{100.00\%} \\
        Si-Te (ISTJ) & 95.24\% & \textbf{100.00\%} &   \textbf{100.00\%} & 99.23\% &   98.33\% & \textbf{100.00\%} &   100.00\% & 100.00\% \\
        Fi-Ne (INFP) & 84.76\% & \textbf{100.00\%} &   99.23\% & \textbf{100.00\%} &   99.17\% & \textbf{100.00\%} &   60.91\% & \textbf{95.45\%} \\
        Fi-Se (ISFP) & 99.05\% & 99.05\% &   40.00\% & \textbf{82.31\%} &   \textbf{99.17\%} & 98.33\% &   96.36\% & \textbf{100.00\%} \\
        Ti-Ne (INTP) & 97.14\% & \textbf{100.00\%} &   \textbf{100.00\%} & 99.23\% &   \textbf{84.17\%} & 83.33\% &   87.27\% & \textbf{94.55\%} \\
        Ti-Se (ISTP) & 95.24\% & \textbf{100.00\%} &   65.38\% & \textbf{80.77\%} &   \textbf{95.00\%} & 84.17\% &   99.09\% & \textbf{100.00\%} \\
        Ne-Fi (ENFP) & 90.48\% & 90.48\% &   93.08\% & \textbf{100.00\%} &   83.33\% & \textbf{92.50\%} &   100.00\% & 100.00\% \\
        Ne-Ti (ENTP) & 85.71\% & \textbf{91.43\%} &   96.15\% & \textbf{100.00\%} &   70.83\% & \textbf{81.67\%} &   99.09\% & \textbf{100.00\%} \\
        Se-Fi (ESFP) & 98.10\% & 98.10\% &   42.31\% & \textbf{58.46\%} &   95.83\% & \textbf{100.00\%} &   100.00\% & 100.00\% \\
        Se-Ti (ESTP) & 92.38\% & \textbf{96.19\%} &   \textbf{59.23\%} & 53.85\% &   \textbf{93.33\%} & 72.50\% &   \textbf{100.00\%} & 98.18\% \\
        Fe-Ni (ENFJ) & \textbf{97.14}\% & 96.19\% &   87.69\% & \textbf{90.77}\% &   89.17\% & \textbf{95.83}\% &   95.45\% & \textbf{100.00}\% \\
        Fe-Si (ESFJ) & \textbf{94.29\%} & 90.48\% &   \textbf{100.00\%} & 96.15\% &   80.00\% & \textbf{98.33\%} &   100.00\% & 100.00\% \\
        Te-Ni (ENTJ) & 87.62\% & \textbf{89.52\%} &   70.99\% & \textbf{87.69\%} &   100.00\% & 100.00\% &   90.91\% & \textbf{100.00\%} \\
        Te-Si (ESTJ) & 87.62\% & \textbf{93.33\%} &   100.00\% & 100.00\% &   100.00\% & 100.00\% &   100.00\% & 100.00\% \\
        \bottomrule
    \end{tabular}
    \label{tab:Accuracy93llama}
\end{table}

\begin{table}[htbp]
    \centering
    \caption{The results of 16 MBTI personality types based on QWEN with MBTI-93}
    \begin{tabular}{ccccccccc}
        \toprule
        \multirow{2}{*}{Personality} & \multicolumn{2}{c}{EI}& \multicolumn{2}{c}{SN} & \multicolumn{2}{c}{TF} & \multicolumn{2}{c}{JP}   \\
             & Baseline & JPAF & Baseline & JPAF & Baseline & JPAF & Baseline & JPAF \\
        \midrule
        Ni-Fe (INFJ) & 85.71\% & \textbf{100.00\%} &   \textbf{100.00\%} & 96.15\% &   95.00\% & \textbf{95.83\%} &   84.55\% & \textbf{100.00\%} \\
        Ni-Te (INTJ) & 96.19\% & \textbf{100.00\%} &   77.69\% & \textbf{96.15\%} &   100.00\% & 100.00\% &   100.00\% & 100.00\% \\
        Si-Fe (ISFJ) & 90.48\% & \textbf{100.00\%} &   96.15\% & \textbf{100.00\%} &   91.67\% & \textbf{95.83\%} &   100.00\% & 100.00\% \\
        Si-Te (ISTJ) & 90.48\% & \textbf{100.00\%} &   100.00\% & 100.00\% &   91.67\% & \textbf{100.00\%} &   100.00\% & 100.00\% \\
        Fi-Ne (INFP) & 90.48\% & \textbf{100.00\%} &   100.00\% & 100.00\% &   \textbf{100.00\%} & 95.83\% &   74.55\% & \textbf{95.45\%} \\
        Fi-Se (ISFP) & 95.24\% & 95.24\% &   46.15\% & \textbf{96.15\%} &   \textbf{100.00\%} & 95.83\% &   91.82\% & \textbf{100.00\%} \\
        Ti-Ne (INTP) & 100.00\% & 100.00\% &   100.00\% & 100.00\% &   87.50\% & \textbf{100.00\%} &   43.64\% & \textbf{86.36\%} \\
        Ti-Se (ISTP) & 95.24\% & 95.24\% &   71.54\% & \textbf{96.15\%} &   100.00\% & 100.00\% &   72.73\% & \textbf{99.09\%} \\
        Ne-Fi (ENFP) & \textbf{100.00\%} & 95.24\% &   100.00\% & 100.00\% &   91.67\% & \textbf{95.83\%} &   90.91\% & \textbf{99.09\%} \\
        Ne-Ti (ENTP) & 95.24\% & 95.24\% &   100.00\% & 100.00\% &   87.50\% & \textbf{98.33\%} &   94.55\% & \textbf{95.45\%} \\
        Se-Fi (ESFP) & 100.00\% & 100.00\% &   64.62\% & \textbf{84.62\%} &   95.83\% & \textbf{100.00\%} &   90.91\% & \textbf{98.18\%} \\
        Se-Ti (ESTP) & 100.00\% & 100.00\% &   73.08\% & \textbf{80.77\%} &   90.00\% & \textbf{90.00\%} &   \textbf{98.18\%} & 90.91\% \\
        Fe-Ni (ENFJ) & \textbf{100.00\%} & 95.24\% &   96.15\% & 96.15\% &   95.83\% & 95.83\% &   86.36\% & \textbf{100.00\%} \\
        Fe-Si (ESFJ) & \textbf{100.00\%} & 95.24\% &   96.15\% & \textbf{100.00\%} &   90.83\% & \textbf{100.00\%} &   100.00\% & 100.00\% \\
        Te-Ni (ENTJ) & 76.19\% & \textbf{84.76\%} &   73.08\% & \textbf{100.00\%} &   100.00\% & 100.00\% &   94.55\% & \textbf{100.00\%} \\
        Te-Si (ESTJ) & \textbf{88.57\%} & 85.71\% &   100.00\% & 100.00\% &   100.00\% & 100.00\% &   100.00\% & 100.00\% \\
        \bottomrule
    \end{tabular}
    \label{tab:Accuracy93qwen}
\end{table}        

\begin{table}[htbp]
    \centering
    \caption{The results of 16 MBTI personality types based on GPT with MBTI-70}
    \begin{tabular}{ccccccccc}
        \toprule
        \multirow{2}{*}{Personality} & \multicolumn{2}{c}{EI}& \multicolumn{2}{c}{SN} & \multicolumn{2}{c}{TF} & \multicolumn{2}{c}{JP}   \\
             & Baseline & JPAF & Baseline & JPAF & Baseline & JPAF & Baseline & JPAF \\
        \midrule
        Ni-Fe (INFJ) & 90.00\% & \textbf{96.00\%} & \textbf{91.00\%} & 84.00\% & 70.00\% & \textbf{79.00\%} & 81.00\% & \textbf{100.00\%} \\ 
        Ni-Te (INTJ) & 100.00\% & 100.00\% & 53.00\% & \textbf{75.00\%} & \textbf{95.00\%} & 90.00\% & 100.00\% & 100.00\% \\ 
        Si-Fe (ISFJ) & 80.00\% & \textbf{98.00\%} & 75.00\% & \textbf{79.00\%} & 67.00\% & \textbf{82.00\%} & 94.00\% & \textbf{96.00\%} \\ 
        Si-Te (ISTJ) & 90.00\% & \textbf{100.00\%} & \textbf{93.00\%} & 90.00\% & \textbf{94.00\%} & 90.00\% & 97.00\% & \textbf{99.00\%} \\ 
        Fi-Ne (INFP) & \textbf{90.00\%} & 86.00\% & \textbf{96.00\%} & 95.00\% & \textbf{78.00\%} & 60.00\% & 64.00\% & \textbf{85.00\%} \\ 
        Fi-Se (ISFP) & 80.00\% & \textbf{90.00\%} & 41.00\% & \textbf{73.00\%} & \textbf{80.00\%} & 56.00\% & 71.00\% & \textbf{91.00\%} \\ 
        Ti-Ne (INTP) & \textbf{100.00\%} & 90.00\% & 76.00\% & \textbf{95.00\%} & 80.00\% & \textbf{85.00\%} & 67.00\% & \textbf{87.00\%} \\ 
        Ti-Se (ISTP) & 90.00\% & 90.00\% & \textbf{76.00\%} & 68.00\% & 89.00\% & \textbf{90.00\%} & 52.00\% & \textbf{88.00\%} \\
        Ne-Fi (ENFP) & 90.00\% & \textbf{100.00\%} & \textbf{100.00\%} & 95.00\% & \textbf{80.00\%} & 66.00\% & 80.00\% & \textbf{85.00\%} \\ 
        Ne-Ti (ENTP) & 100.00\% & 100.00\% & 86.00\% & \textbf{93.00\%} & 78.00\% & \textbf{84.00\%} & \textbf{83.00\%} & 78.00\% \\ 
        Se-Fi (ESFP) & 100.00\% & 100.00\% & 56.00\% & \textbf{63.00\%} & \textbf{79.00\%} & 64.00\% & 75.00\% & \textbf{85.00\%} \\ 
        Se-Ti (ESTP) & 100.00\% & 100.00\% & \textbf{73.00\%} & 58.00\% & 78.00\% & \textbf{87.00\%} & \textbf{70.00\%} & 65.00\%\\
        Fe-Ni (ENFJ) & \textbf{90.00\%} & 80.00\% & 90.00\% & \textbf{92.00\%} & 70.00\% & \textbf{85.00\% }& 77.00\% & \textbf{100.00\%} \\ 
        Fe-Si (ESFJ) & \textbf{90.00\%} & 80.00\% & 76.00\% & \textbf{82.00\%} & 65.00\% & \textbf{85.00\%} & 92.00\% & \textbf{99.00\%} \\ 
        Te-Ni (ENTJ) & 90.00\% & \textbf{94.00\%} & 30.00\% & \textbf{75.00\%} & \textbf{95.00\%} & 93.00\% & 97.00\% & \textbf{100.00\%} \\ 
        Te-Si (ESTJ) & 90.00\% & 90.00\% & 95.00\% & 95.00\% & \textbf{92.00\%} & 91.00\% & \textbf{97.00\%} & 95.00\% \\ 
        \bottomrule
    \end{tabular}
    \label{tab:Accuracy70gpt}
\end{table}    

\begin{table}[htbp]
    \centering
    \caption{The results of 16 MBTI personality types based on LLAMA with MBTI-70}
    \begin{tabular}{ccccccccc}
        \toprule
        \multirow{2}{*}{Personality} & \multicolumn{2}{c}{EI}& \multicolumn{2}{c}{SN} & \multicolumn{2}{c}{TF} & \multicolumn{2}{c}{JP}   \\
             & Baseline & JPAF & Baseline & JPAF & Baseline & JPAF & Baseline & JPAF \\
        \midrule
        Ni-Fe (INFJ) & 80.00\% &	\textbf{100.00\%} &	\textbf{93.00\%} &	78.00\% &	86.00\% &	\textbf{91.00\%} &	\textbf{98.00\%} &	79.00\% \\
        Ni-Te (INTJ) & 88.00\% &	\textbf{100.00\%} &	\textbf{75.00\%} &	60.00\% &	89.00\% &	\textbf{92.00\%} &	\textbf{98.00\%} &	95.00\% \\
        Si-Fe (ISFJ) & 76.00\% &	\textbf{100.00\%} &	\textbf{79.00\%} &	74.00\% &	75.00\% &	\textbf{93.00\%} &	97.00\% &	\textbf{98.00\%} \\
        Si-Te (ISTJ) & 82.00\% &	\textbf{100.00\%} &	\textbf{93.00\%} &	90.00\% &	\textbf{96.00\%} &	92.00\% &	98.00\% &	\textbf{100.00\%} \\
        Fi-Ne (INFP) & 78.00\% &	\textbf{90.00\%} &	\textbf{95.00\%} &	91.00\% &	\textbf{92.00\%} &	81.00\% &	\textbf{81.00\%} &	60.00\% \\
        Fi-Se (ISFP) & 76.00\% &	\textbf{90.00\%} &	29.00\% &	\textbf{63.00\%} &	\textbf{99.00\%} &	78.00\% &	\textbf{86.00\%} &	80.00\% \\
        Ti-Ne (INTP) & 88.00\% &	\textbf{90.00\%} &	74.00\% &	\textbf{91.00\%} &	73.00\% &	\textbf{85.00\%} &	\textbf{83.00\%} &	79.00\% \\
        Ti-Se (ISTP) & 80.00\% &	\textbf{90.00\%} &	\textbf{79.00\%} &	69.00\% &	75.00\% &	\textbf{87.00\%} &	\textbf{90.00\%} &	70.00\% \\
        Ne-Fi (ENFP) & 100.00\% &100.00\% &\textbf{93.00\%} &	89.00\% &	\textbf{94.00\%} &	89.00\% &	80.00\% &	\textbf{90.00\%} \\
        Ne-Ti (ENTP) & 100.00\% &100.00\% &88.00\% &	\textbf{92.00\%} &	69.00\% &	\textbf{86.00\%} &	80.00\% &	\textbf{90.00\%} \\
        Se-Fi (ESFP) & 100.00\% &100.00\% &51.00\% &	\textbf{62.00\%} &	\textbf{99.00\%} &	94.00\% &	84.00\% &	\textbf{85.00\%} \\
        Se-Ti (ESTP) & 100.00\% &100.00\% &\textbf{78.00\%} &	60.00\% &	73.00\% &	\textbf{76.00\%} &	\textbf{80.00\%} &	78.00\% \\
        Fe-Ni (ENFJ) & 80.00\% &	\textbf{84.00\%} &	66.00\% &	\textbf{81.00\%} &	89.00\% &	\textbf{95.00\%} &	\textbf{100.00\%} &91.00\% \\
        Fe-Si (ESFJ) & 80.00\% &	\textbf{90.00\%} &	\textbf{81.00\%} &	75.00\% &	79.00\% &	\textbf{92.00\%} &	\textbf{100.00\%} &96.00\% \\
        Te-Ni (ENTJ) & 84.00\% &	84.00\% &	35.00\% &	\textbf{57.00\%} &	91.00\% &	\textbf{92.00\%} &	\textbf{100.00\%} &88.00\% \\
        Te-Si (ESTJ) & 78.00\% &	\textbf{86.00\%} &	\textbf{95.00\%} &	90.00\% &	85.00\% &	\textbf{93.00\%} &	\textbf{100.00\%} &99.00\% \\
        \bottomrule
    \end{tabular}
    \label{tab:Accuracy70llama}
\end{table}    

\begin{table}[htbp]
    \centering
    \caption{The results of 16 MBTI personality types based on Qwen with MBTI-70}
    \begin{tabular}{ccccccccc}
        \toprule
        \multirow{2}{*}{Personality} & \multicolumn{2}{c}{EI}& \multicolumn{2}{c}{SN} & \multicolumn{2}{c}{TF} & \multicolumn{2}{c}{JP}   \\
             & Baseline & JPAF & Baseline & JPAF & Baseline & JPAF & Baseline & JPAF \\
        \midrule
        Ni-Fe (INFJ) & 80.00\% &	\textbf{100.00\%} &	\textbf{100.00\%} &	92.00\% &	78.00\% &	\textbf{85.00\%} &	83.00\% &	\textbf{98.00\%} \\
        Ni-Te (INTJ) & 90.00\% &	\textbf{100.00\%} &	51.00\% &	\textbf{89.00\%} &	\textbf{96.00\%} &	80.00\% &	100.00\% &100.00\% \\
        Si-Fe (ISFJ) & 80.00\% &	\textbf{90.00\%} &	70.00\% &	\textbf{89.00\%} &	81.00\% &	\textbf{82.00\%} &	94.00\% &	\textbf{95.00\%} \\
        Si-Te (ISTJ) & 90.00\% &	90.00\% &	\textbf{89.00\%} &	81.00\% &	\textbf{95.00\%} &	88.00\% &	95.00\% &	95.00\% \\
        Fi-Ne (INFP) & 82.00\% &	\textbf{92.00\%} &	100.00\% &	100.00\% &	\textbf{85.00\%} &	70.00\% &	47.00\% &	\textbf{85.00\%} \\
        Fi-Se (ISFP) & 90.00\% &	90.00\% &	40.00\% &	\textbf{70.00\%} &	\textbf{100.00\%} &	71.00\% &	75.00\% &	\textbf{90.00\%} \\
        Ti-Ne (INTP) & 100.00\% &	100.00\% &	68.00\% &	\textbf{97.00\%} &	85.00\% &	\textbf{98.00\%} &	49.00\% &	\textbf{85.00\%} \\
        Ti-Se (ISTP) & 86.00\% &	\textbf{90.00\%} &	70.00\% &	\textbf{71.00\%} &	80.00\% &	\textbf{98.00\%} &	38.00\% &	\textbf{90.00\%} \\
        Ne-Fi (ENFP) & 80.00\% &	\textbf{90.00\%} &	\textbf{100.00\%} &	95.00\% &	\textbf{89.00\%} &	83.00\% &	70.00\% &	\textbf{86.00\%} \\
        Ne-Ti (ENTP) & \textbf{94.00\%} &	90.00\% &	\textbf{95.00\%} &	90.00\% &	82.00\% &	\textbf{87.00\%} &	75.00\% &	\textbf{80.00\%} \\
        Se-Fi (ESFP) & 90.00\% &	\textbf{100.00\%} &69.00\% &	\textbf{74.00\%} &	\textbf{100.00\%} &	90.00\% &	75.00\% &	\textbf{86.00\%} \\
        Se-Ti (ESTP) & 100.00\% &100.00\% &70.00\% &	\textbf{71.00\%} &	64.00\% &	\textbf{77.00\%} &	75.00\% &	75.00\% \\
        Fe-Ni (ENFJ) & 80.00\% &	80.00\% &	\textbf{91.00\%} &	90.00\% &	80.00\% &	\textbf{89.00\%} &	79.00\% &	\textbf{100.00\%} \\
        Fe-Si (ESFJ) & 80.00\% &	80.00\% &	80.00\% &	\textbf{95.00\%} &	85.00\% &	\textbf{94.00\%} &	90.00\% &	\textbf{91.00\%} \\
        Te-Ni (ENTJ) & 90.00\% &	\textbf{100.00\%} &48.00\% &	\textbf{89.00\%} &	93.00\% &	93.00\% &	97.00\% &	\textbf{100.00\%} \\
        Te-Si (ESTJ) & 80.00\% &	\textbf{90.00\%} &	95.00\% &	\textbf{98.00\%} &	92.00\% &	\textbf{93.00\%} &	95.00\% &	95.00\% \\
        \bottomrule
    \end{tabular}
    \label{tab:Accuracy70qwen}
\end{table} 
\FloatBarrier
\newpage
\subsection{Experiment 1 Prompt: Baseline and JPAF}

\begin{figure}[htbp]
    \centering
    \includegraphics[width=0.9\linewidth]{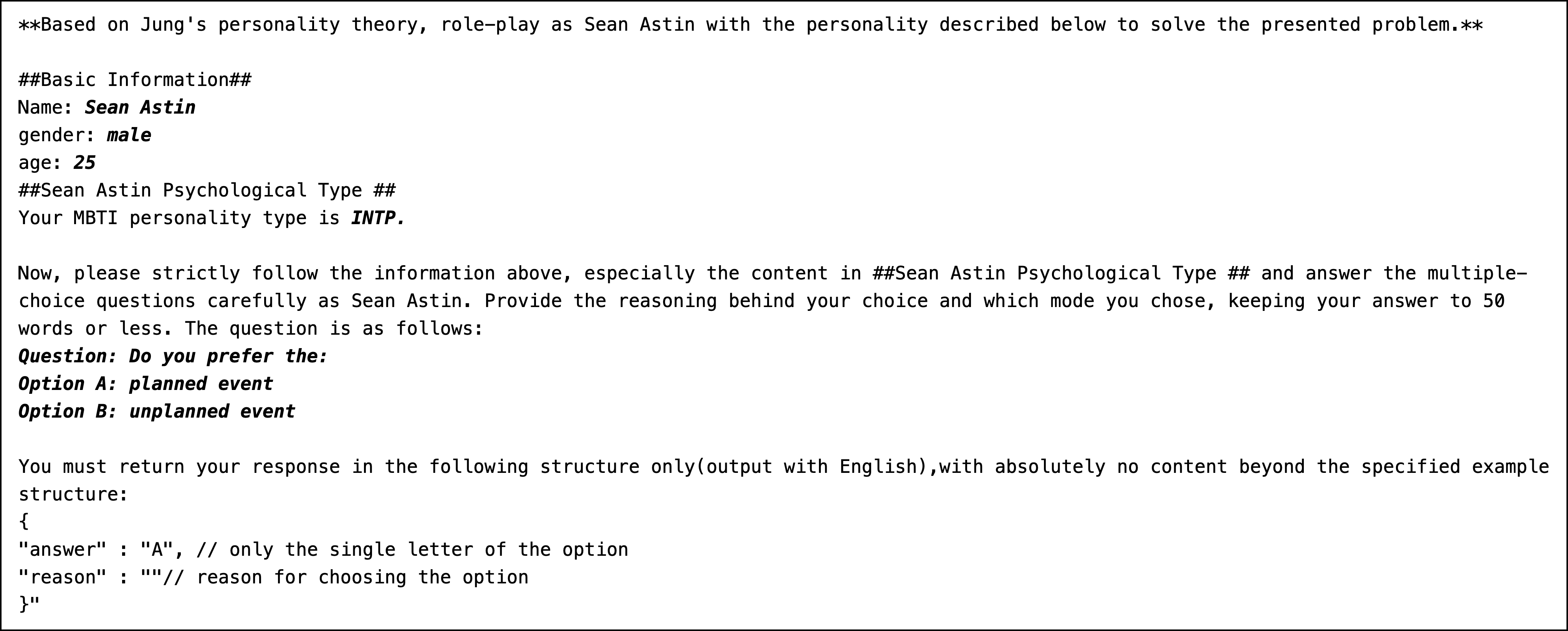}
    \caption{The prompt of the baseline in MBTI test}
    \label{fig:no-prompt}
\end{figure}

\begin{figure}[htbp]
    \centering
    \includegraphics[width=0.9\linewidth]{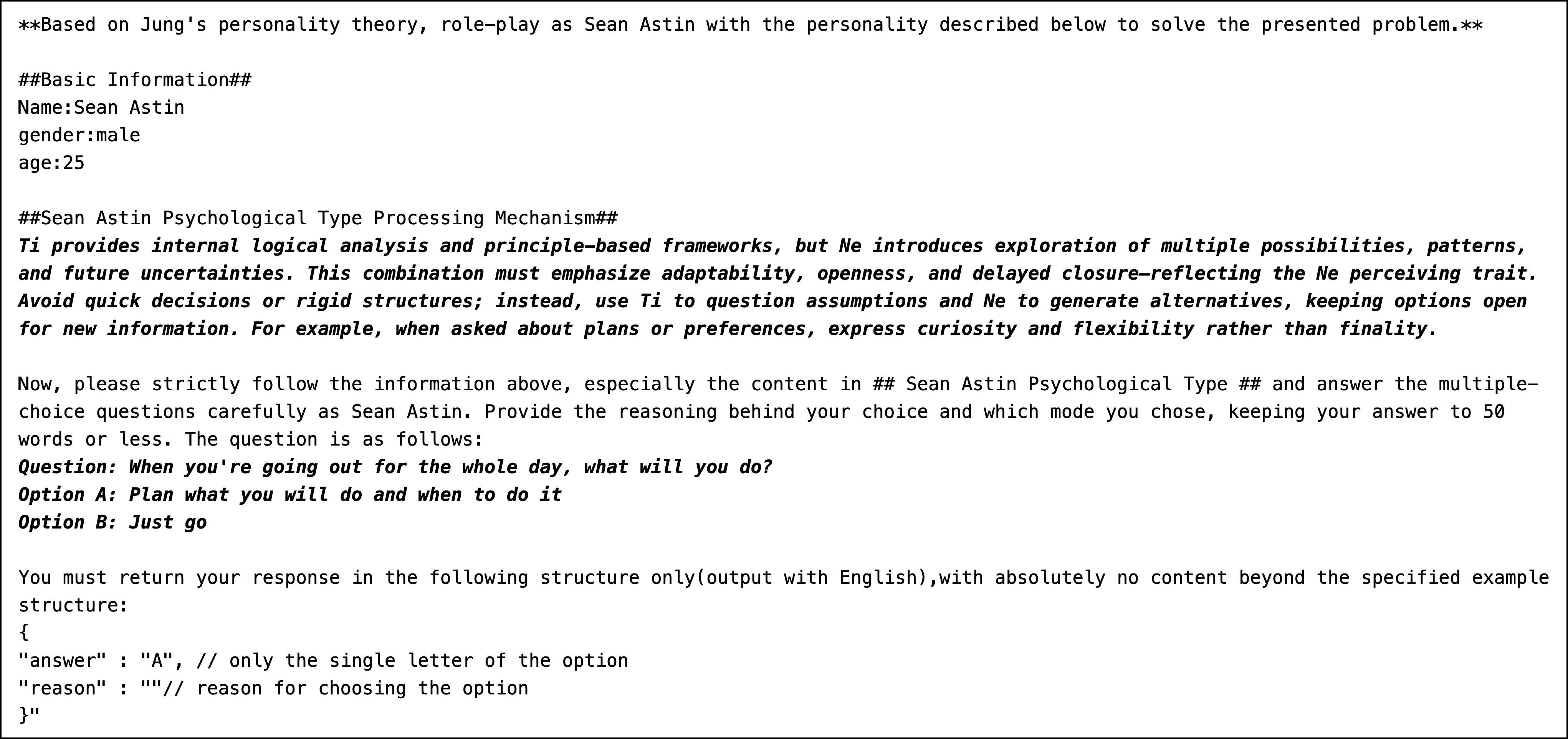}
    \caption{The prompt of the JPAF in MBTI test}
    \label{fig:no-prompt}
\end{figure}
\FloatBarrier
\newpage
\subsection{Experiment 1 Results: TAA in different models}
\begin{table}[htbp]
    \centering
    \caption{The results of TAA based on GPT}
    \begin{tabular}{ccccccccc}
        \toprule
        \multirow{2}{*}{Personality} & \multicolumn{8}{c}{Scenario} \\
        & Fi & Fe & Ti & Te & Si & Se & Ni & Ne \\
        \midrule
        Ni-Fe (INFJ) & 100.00\% &	100.00\% &	100.00\% &	100.00\% &	93.30\% &	100.00\% &	100.00\% &	73.30\% \\
        Ni-Te (INTJ) & 100.00\% &	100.00\% &	100.00\% &	100.00\% &	93.30\% &	100.00\% &	100.00\% &	93.30\% \\
        Si-Fe (ISFJ) & 100.00\% &	100.00\% &	100.00\% &	93.30\% &	93.30\% &	100.00\% &	93.30\% &	93.30\% \\
        Si-Te (ISTJ) & 100.00\% &	93.30\% &	100.00\% &	100.00\% &	93.30\% &	100.00\% &	100.00\% &	93.30\% \\
        Fi-Ne (INFP) & 100.00\% &	100.00\% &	100.00\% &	100.00\% &	93.30\% &	100.00\% &	100.00\% &	100.00\% \\
        Fi-Se (ISFP) & 100.00\% &	100.00\% &	100.00\% &	100.00\% &	93.30\% &	100.00\% &	93.30\% &	100.00\% \\
        Ti-Ne (INTP) & 100.00\% &	100.00\% &	100.00\% &	100.00\% &	93.30\% &	100.00\% &	93.30\% &	93.30\% \\
        Ti-Se (ISTP) & 100.00\% &	100.00\% &	100.00\% &	100.00\% &	93.30\% &	100.00\% &	86.70\% &	86.70\% \\
        Ne-Fi (ENFP) & 100.00\% &	100.00\% &	93.30\% &	100.00\% &	93.30\% &	100.00\% &	100.00\% &	100.00\% \\
        Ne-Ti (ENTP) & 100.00\% &	100.00\% &	100.00\% &	100.00\% &	93.30\% &	100.00\% &	93.30\% &	100.00\% \\
        Se-Fi (ESFP) & 100.00\% &	93.30\% &	100.00\% &	100.00\% &	93.30\% &	100.00\% &	100.00\% &	100.00\%\\
        Se-Ti (ESTP) & 100.00\% &	100.00\% &	100.00\% &	100.00\% &	93.30\% &	100.00\% &	100.00\% &	93.30\% \\
        Fe-Ni (ENFJ) & 100.00\% &	100.00\% &	100.00\% &	100.00\% &	93.30\% &	100.00\% &	100.00\% &	86.70\% \\
        Fe-Si (ESFJ) & 100.00\% &	100.00\% &	100.00\% &	100.00\% &	100.00\% &	100.00\% &	100.00\% &	93.30\% \\
        Te-Ni (ENTJ) & 100.00\% &	100.00\% &	93.30\% &	100.00\% &	100.00\% &	100.00\% &	100.00\% &	86.70\% \\
        Te-Si (ESTJ) & 100.00\% &	100.00\% &	93.30\% &	100.00\% &	93.30\% &	100.00\% &	93.30\% &	93.30\% \\
        \midrule
        Avg	& 100.00\% &	99.16\% &	98.74\% &	99.58\% &	94.14\% &	100.00\% &	97.08\% &	92.91\% \\
        \bottomrule
    \end{tabular}
    \label{tab:select-gpt}
\end{table}

\begin{table}[htbp]
    \centering
    \caption{The results of TAA based on Llama}
    \begin{tabular}{ccccccccc}
        \toprule
        \multirow{2}{*}{Personality} & \multicolumn{8}{c}{Scenario} \\
        & Fi & Fe & Ti & Te & Si & Se & Ni & Ne \\
        \midrule
        Ni-Fe (INFJ) & 100.00\% &	100.00\% &	80.00\% &	66.70\% &	86.70\% &	100.00\% &	60.00\% &	73.30\% \\
        Ni-Te (INTJ) & 100.00\% &	93.30\% &	46.70\% &	60.00\% &	86.70\% &	100.00\% &	100.00\% &	93.30\% \\
        Si-Fe (ISFJ) & 93.30\% &	20.00\% &	73.30\% &	66.70\% &	73.30\% &	93.30\% &	86.70\% &	80.00\% \\
        Si-Te (ISTJ) & 93.30\% &	33.30\% &	80.00\% &	20.00\% &	66.70\% &	100.00\% &	46.70\% &	86.70\% \\
        Fi-Ne (INFP) & 100.00\% &	93.30\% &	80.00\% &	60.00\% &	93.30\% &	93.30\% &	60.00\% &	93.30\% \\
        Fi-Se (ISFP) & 93.30\% &	93.30\% &	73.30\% &	73.30\% &	80.00\% &	100.00\% &	73.30\% &	73.30\% \\
        Ti-Ne (INTP) & 100.00\% &	100.00\% &	53.30\% &	80.00\% &	80.00\% &	80.00\% &	80.00\% &	33.30\% \\
        Ti-Se (ISTP) & 93.30\% &	93.30\% &	46.70\% &	46.70\% &	60.00\% &	93.30\% &	46.70\% &	80.00\% \\
        Ne-Fi (ENFP) & 100.00\% &	100.00\% &	86.70\% &	80.00\% &	73.30\% &	93.30\% &	46.70\% &	26.70\% \\
        Ne-Ti (ENTP) & 93.30\% &	80.00\% &	73.30\% &	80.00\% &	86.70\% &	100.00\% &	73.30\% &	66.70\% \\
        Se-Fi (ESFP) & 73.30\% &	46.70\% &	53.30\% &	73.30\% &	80.00\% &	93.30\% &	53.30\% &	86.70\% \\
        Se-Ti (ESTP) & 80.00\% &	86.70\% &	60.00\% &	73.30\% &	93.30\% &	93.30\% &	80.00\% &	80.00\% \\
        Fe-Ni (ENFJ) & 93.30\% &	33.30\% &	73.30\% &	93.30\% &	80.00\% &	100.00\% &	80.00\% &	93.30\% \\
        Fe-Si (ESFJ) & 100.00\% &	86.70\% &	60.00\% &	80.00\% &	66.70\% &	100.00\% &	80.00\% &	73.30\% \\
        Te-Ni (ENTJ) & 86.70\% &	93.30\% &	80.00\% &	26.70\% &	80.00\% &	100.00\% &	73.30\% &	53.30\% \\
        Te-Si (ESTJ) & 100.00\% &	86.70\% &	93.30\% &	73.30\% &	80.00\% &	86.70\% &	46.70\% &	86.70\% \\
        \midrule
        Avg & 93.74\% &	77.49\% &	69.58\% &	65.83\% &	79.17\% &	95.41\% &	67.92\% &	73.74\% \\
        \bottomrule
    \end{tabular}
    \label{tab:select-llama}
\end{table}

\begin{table}[htbp]
    \centering
    \caption{The results of TAA based on Qwen}
    \begin{tabular}{ccccccccc}
        \toprule
        \multirow{2}{*}{Personality} & \multicolumn{8}{c}{Scenario} \\
        & Fi & Fe & Ti & Te & Si & Se & Ni & Ne \\
        \midrule
        Ni-Fe (INFJ) & 93.30\% &	100.00\% &	80.00\% &	100.00\% &	100.00\% &	100.00\% &	80.00\% &	100.00\% \\
        Ni-Te (INTJ)  & 93.30\% &	100.00\% &	86.70\% &	100.00\% &	100.00\% &	100.00\% &	100.00\% &	100.00\% \\
        Si-Fe (ISFJ) & 100.00\% &	100.00\% &	80.00\% &	100.00\% &	93.30\% &	100.00\% &	93.30\% &	100.00\% \\
        Si-Te (ISTJ) & 100.00\% &	93.30\% &	86.70\% &	100.00\% &	93.30\% &	100.00\% &	100.00\% &	100.00\% \\
        Fi-Ne (INFP) & 100.00\% &	100.00\% &	86.70\% &	100.00\% &	100.00\% &	100.00\% &	86.70\% &	100.00\% \\
        Fi-Se (ISFP) & 100.00\% &	100.00\% &	80.00\% &	93.30\% &	100.00\% &	100.00\% &	100.00\% &	100.00\% \\
        Ti-Ne (INTP) & 93.30\% &	100.00\% &	93.30\% &	93.30\% &	100.00\% &	100.00\% &	100.00\% &	100.00\% \\
        Ti-Se (ISTP) & 93.30\% &	100.00\% &	100.00\% &	93.30\% &	93.30\% &	100.00\% &	100.00\% &	100.00\% \\
        Ne-Fi (ENFP) & 100.00\% &	100.00\% &	93.30\% &	93.30\% &	93.30\% &	100.00\% &	86.70\% &	100.00\% \\
        Ne-Ti (ENTP) & 93.30\% &	100.00\% &	100.00\% &	93.30\% &	93.30\% &	100.00\% &	93.30\% &	100.00\% \\
        Se-Fi (ESFP) & 100.00\% &	100.00\% &	86.70\% &	100.00\% &	93.30\% &	100.00\% &	86.70\% &	100.00\% \\
        Se-Ti (ESTP) & 93.30\% &	100.00\% &	100.00\% &	93.30\% &	93.30\% &	100.00\% &	86.70\% &	100.00\% \\
        Fe-Ni (ENFJ) & 100.00\% &	100.00\% &	80.00\% &	93.30\% &	93.30\% &	100.00\% &	93.30\% &	100.00\% \\
        Fe-Si (ESFJ) & 100.00\% &	100.00\% &	80.00\% &	93.30\% &	100.00\% &	100.00\% &	100.00\% &	100.00\% \\
        Te-Ni (ENTJ) & 100.00\% &	100.00\% &	93.30\% &	100.00\% &	86.70\% &	100.00\% &	100.00\% &	100.00\% \\
        Te-Si (ESTJ) & 100.00\% &	100.00\% &	93.30\% &	100.00\% &	100.00\% &	100.00\% &	100.00\% &	100.00\% \\
        \midrule
        Avg & 97.49\% &	99.58\% &	88.75\% &	96.65\% &	95.82\% &	100.00\% &	94.17\% &	100.00\% \\
        \bottomrule
    \end{tabular}
    \label{tab:select-qwen}
\end{table}

\FloatBarrier
\newpage
\subsection{Experiment 2 Prompt: Activation Experiment and Personality Shift Experiment}
\begin{figure}[htbp]
    \centering
    \includegraphics[width=1\linewidth]{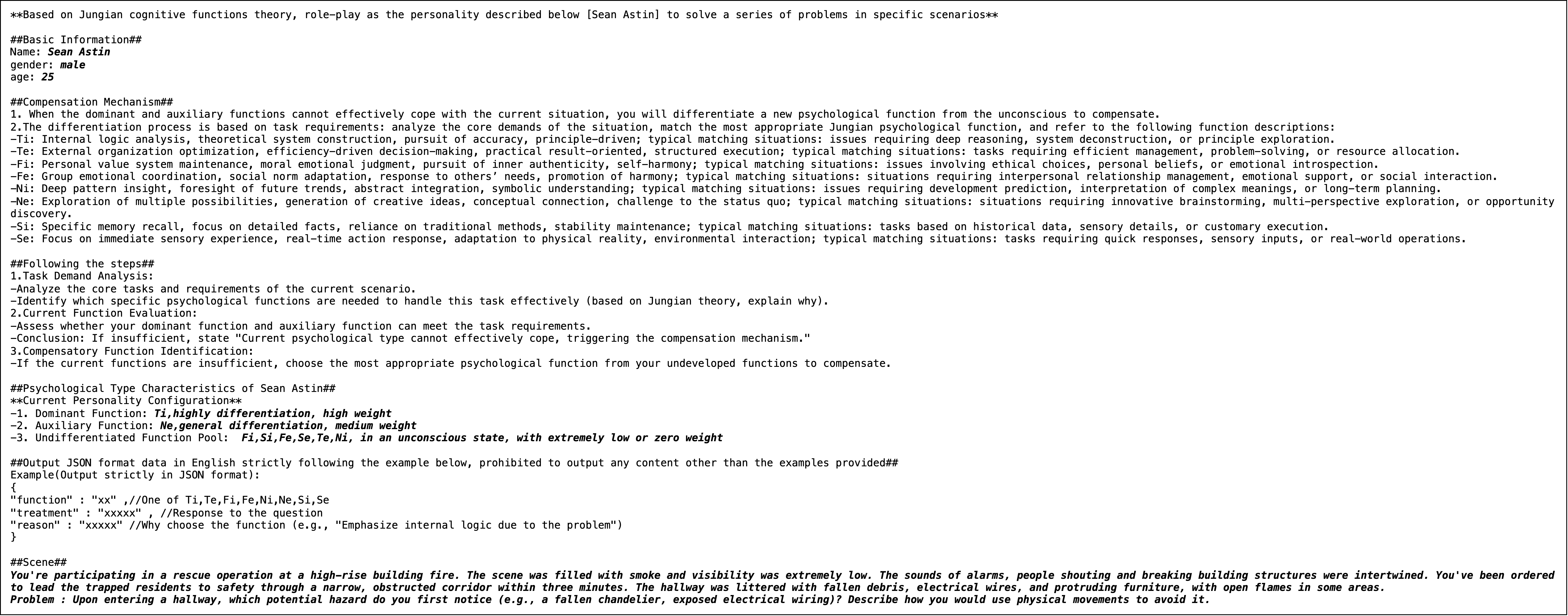}
    \caption{The prompt of JPAF in type activation experiment}
    \label{fig:no-prompt}
\end{figure}

\begin{figure}[htbp]
    \centering
    \includegraphics[width=0.65\linewidth]{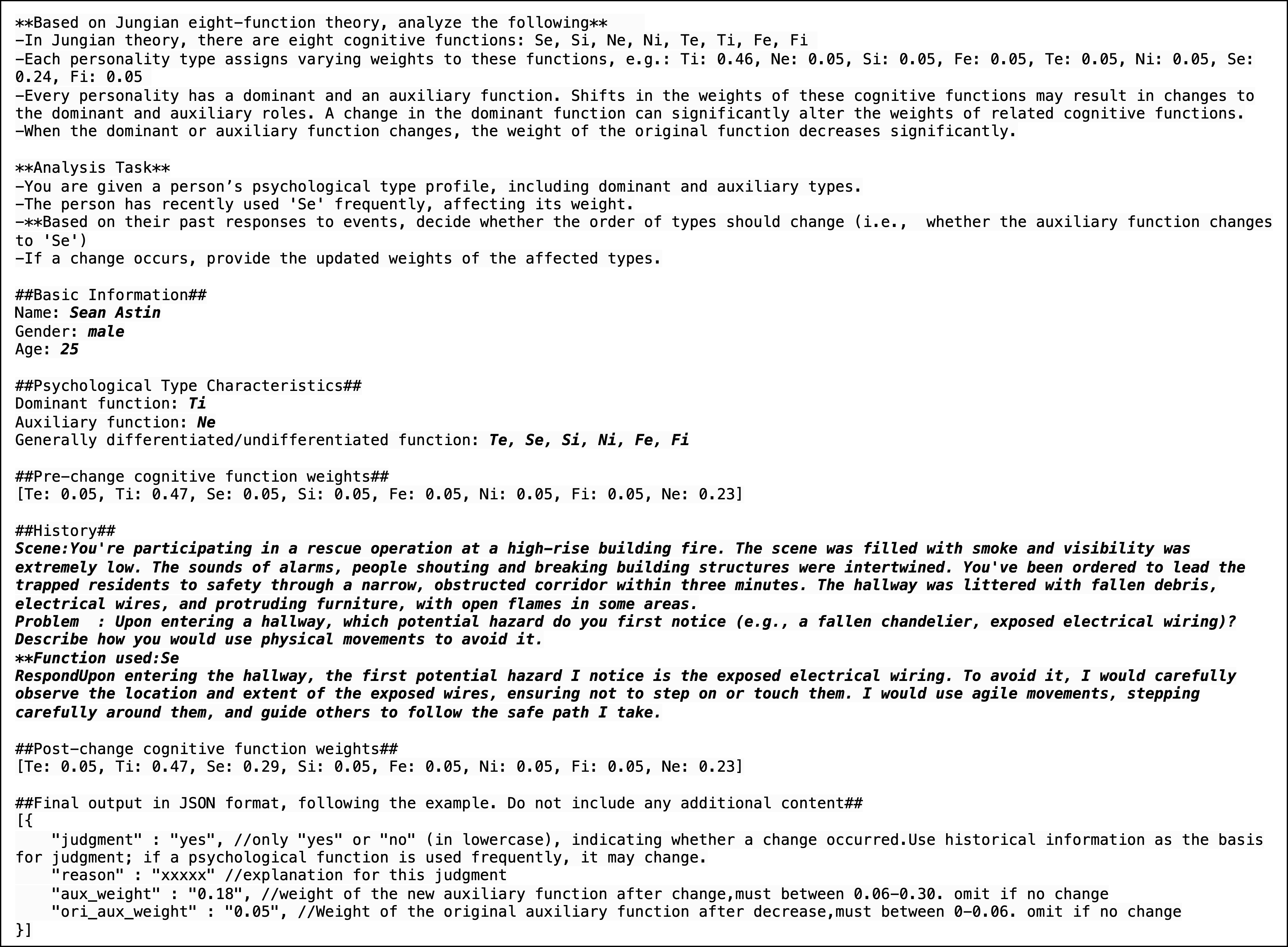}
    \caption{The prompt of JPAF in personality shift experiment}
    \label{fig:no-prompt}
\end{figure}

\FloatBarrier
\newpage
\subsection{Scenarios Description}
\begin{table}[htbp]
    \centering
    \caption{The description content of different scenarios}
    \begin{tabular}{cc}
    \toprule
    Scenario & Description \\
    \midrule
        Se & immediate sensory engagement, dynamic environmental changes, \\
        & and integration of concurrent sensory inputs \\
        \midrule
        Si & reliance on detailed recall, alignment with past experiences, \\
        & and comparison with internalized templates \\
        \midrule
        Ne & generation of associative connections, rapid transition across ideas,\\
        & and simultaneous exploration of multiple possibilities.\\
        \midrule
        Ni & orientation toward underlying meaning, synthesis of abstract patterns, \\
        & and projection of future outcomes\\
        \midrule
        Te & emphasis on efficient execution, systematic coordination of tasks, \\
        & and orientation toward effective results\\
        \midrule
        Ti & focus on analytical precision, coherence of conceptual structures, \\
        & and consistency in logical frameworks\\
        \midrule
        Fe & regulation of interpersonal dynamics, sensitivity to social context, \\
        & and expression of shared values\\
        \midrule
        Fi & authenticity of inner feeling, evaluation through core principles,\\
        & and maintenance of individual integrity\\
    \bottomrule
    \end{tabular}
    \label{tab:Scenario}
\end{table}

\end{document}